\documentclass{article}
\ifdefined\pdfoutput\pdfoutput=1\fi
\usepackage{arxiv}
\usepackage{iftex}
\ifPDFTeX
	\usepackage[T1]{fontenc}
\else
	\usepackage[no-math]{fontspec}
\fi
\usepackage{amsmath,amssymb}
\usepackage{graphicx}
\usepackage{booktabs}
\usepackage{array}
\usepackage[numbers,sort&compress]{natbib}
\usepackage{hyperref}
\hypersetup{
	hidelinks,
	linktoc = all,
	pdfdisplaydoctitle,
	breaklinks,
	pdfstartview = Fit,
	unicode,
	pdftitle={Matched-null testing of word embedding geometry},
	pdfsubject={word embeddings, geometric data analysis, null models},
	pdfauthor={Daniel Quigley},
	pdfkeywords={word embeddings, null models, intrinsic dimension, Ollivier-Ricci curvature, Gromov hyperbolicity, surrogate data},
}
\usepackage[activate={true,nocompatibility},final]{microtype}
\usepackage{doi}
\newcommand{\tabhead}[1]{{\boldmath\bfseries #1}}

\title{Matched-null testing of word embedding geometry}

\author{ \href{https://orcid.org/0009-0004-7957-1806}{\includegraphics[scale=0.06]{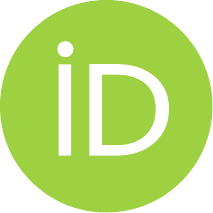}\hspace{1mm}Daniel Quigley} \\
	Center for Possible Minds\\
	Indiana University Bloomington\\
	Bloomington, IN 47408 \\
	\texttt{dgquigle@iu.edu} \\
}

\renewcommand{\headeright}{Preprint}
\renewcommand{\undertitle}{Preprint}
\renewcommand{\shorttitle}{Matched-null testing}

\begin{document}
\maketitle

\begin{abstract}
    This paper represents an on-going statistical investigation into the properties and shape of semantic space, in which we probe for structure in embeddings. On Gaussian samples with a word embedding's covariance, statistics of intrinsic dimension, curvature, and hyperbolicity take values that would count as evidence of structure in an embedding; we argue that an embedding's value supports a structural claim only when it departs from the statistic's distribution over these samples, the covariance-matched null. Crediting the structure to the training corpus requires a second comparison, against embeddings retrained on shuffled text. Matched-null testing reports each difference in standard deviations of its null. On the unit sphere, with 2000-word samples and 20 neighbors, the Levina--Bickel estimator gives $26$ to $27$ for GloVe and Word2Vec, against $61$ to $71$ for their nulls, which span the full space. Skip-gram and GloVe trained on a Wikipedia slice show a similar deficit, and retrained on shuffled copies they retain at most $13\%$ of it. Since shuffling keeps each word's frequency and randomizes its contexts, the deficit depends mostly on co-occurrence. Every model trained on intact text exceeds its null in mean Ollivier--Ricci curvature, yet skip-gram retrained on shuffled text reproduces $91\%$ of that excess through hubs. On intact text, skip-gram's graph has few hubs, so the source of its excess is undetermined. Khrulkov and colleagues read a low ratio of Gromov $\delta$ to diameter as evidence that data suit hyperbolic embeddings. On the sphere, the ratio is within two standard deviations of its null mean.
\end{abstract}

\keywords{word embeddings \and null models \and intrinsic dimension \and Ollivier--Ricci curvature \and Gromov hyperbolicity \and surrogate data}

\section{Introduction}\label{sec:intro}

	Much may be said about the geometric character of a semantic space, of which geometric statistics are used to characterize such word embeddings: \citet{cai2020} report low local intrinsic dimension in contextual embeddings; \citet{ethayarajh2019} measures their anisotropy; \citet{jakubowski2020} use persistent homology to investigate polysemy. \citet{khrulkov2020} define a relative hyperbolicity $2\delta/\mathrm{diameter}$ and take low values as evidence that a dataset suits hyperbolic embeddings. Forman curvature, Ollivier curvature, and the $k$NN degree distribution tested here are our formulations; on Gaussian samples with an embedding's covariance, estimators of dimension, curvature, hyperbolicity, and degree already return values as structure, through distance concentration, high-dimensional $k$NN hubness \citep{Radovanovi2010}, and the anisotropy of the covariance. Here, we are interested in the conditions under which a value of such a statistic supports structure.

	\citet{theiler1992} address a similar problem in time-series analysis, by comparing dimension estimates with those of surrogate data preserving linear correlations; we adapt surrogate-data testing to embedding geometry; we compare each statistic with its distribution over draws from a null, and its difference from the null mean in the statistic's units and in units of the null's standard deviation. The isotropic null, a standard Gaussian, maintains only the sample size and the dimension. The covariance-matched null, a Gaussian fitted to the embedding's mean and covariance, maintains its second moments. The marginal-permuted null, which permutes each coordinate across rows, maintains every univariate marginal. Here, tracking the covariance-matched null is indistinguishable from a functional of the embedding's second moments. For embeddings used with cosine similarity, we normalize the rows to the unit sphere, project the null draws onto it, and compare the participation ratio between the projected null and the data.

	We apply a suite of statistics to classic word embeddings of GloVe \citep{pennington2014glove}, trained on a subset of Dolma \citep{dolma2024} and released by \citet{carlson2025}, and to Word2Vec, trained on Google News \citep{mikolov2013}. On the unit sphere, both embeddings have lower local intrinsic dimension and greater neighborhood overlap, measured by Ollivier--Ricci curvature, than their covariance-matched nulls. Forman--Ricci curvature the same overlap through its triangle term, and its difference from the null against the hub-and-bridge reading. The ratio $\delta/\mathrm{diameter}$ and the maximum $k$NN degree track their nulls. On the unnormalized GloVe rows, these two findings reverse, and the dimension and Ollivier findings persist.

	Following \citet{valentini2022,valentini2023}, we train skip-gram and GloVe on a Wikipedia slice and on three copies of it whose tokens are permuted at random within blocks of $10^8$ tokens, and so each word maintains frequency and loses context. Under shuffling, both configurations lose most of their local dimension deficit; the deficit, therefore, depends largely on co-occurrence. GloVe's curvature excess also passes the comparison, since its shuffled models fall below their own null. That null is a thin Gaussian, unlike the original's. The attribution consequently holds only as a bound. The source of skip-gram's excess stays undetermined: shuffled skip-gram reproduces $91\%$ of it through hubs, with twelve times the maximum degree of the intact-corpus model.

	Both surviving signals persist after we remove the densest clusters and at the scale of cluster centroids (where the empirical null applies). In the released embeddings, the tightest clusters are lists from recurring corpus formats, as in news boilerplate and athlete rosters, or as frequency bands; in the Wikipedia-trained models, two of eight clusters read as topical classes.

	On the sphere, the null approximately matches GloVe's second moments. For Word2Vec it misses the participation ratio by $13\sigma$, in the conservative direction for both signals. Results within about $7\sigma$ of zero stay provisional, because a replicate draw puts two subsets of one condition three null standard deviations apart. For the released and the intact-corpus models, $\delta/\mathrm{diameter}$ is below its null mean in every point-level and cluster-removed condition. Across the centroid clouds, its sign varies. A pooled test over independent subsets would decide whether the negative sign marks tree-like structure. The corpus comparison rests on one Wikipedia slice, one trained model per configuration, and three shuffles.

    This paper is an on-going program into probing for structure in semantic spaces. Presently, the paper concerns static embeddings; future work will account for contextual embeddings and as embeddings change over the course of training and application. Collaboration is invited and welcome.

\section{Matched-null testing}\label{sec:method}
    
	Matched-null testing is a statistic computed on the cloud as its standardized excess over draws from a null that preserves part of the cloud's structure.

\subsection{Setup}
	Let $X \in \mathbb{R}^{n \times d}$ be a point cloud and $T(\cdot)$ a geometric statistic. The matched-null test reports the raw excess $T(X) - \mathbb{E}[T(Z)]$ with its standardized form
	\begin{equation}
		\sigma_T(X; Z) = \frac{T(X) - \mathbb{E}[T(Z)]}{\sqrt{\mathrm{Var}[T(Z)]}},
		\label{eq:excess}
	\end{equation}
	where $Z$ is a null distribution constructed from $X$ in such a way that it preserves a specified subset of $X$'s structure, and destroys the rest. In practice, we estimate $\mathbb{E}[T(Z)]$ and $\mathrm{Var}[T(Z)]$ from $K$ independent draws $\{Z^{(k)}\}_{k=1}^{K}$, and take the empirical analog $(T(X) - \bar T(Z))/s(Z)$, with $\bar T(Z) = \frac{1}{K}\sum_k T(Z^{(k)})$ and $s(Z)$ the sample standard deviation. Every null uses $K = 10$ draws except the per-cluster nulls, which use $K = 5$. Because $s(Z)$ is, itself, estimated, comparisons of $\sigma_T$ across conditions are only as stable as $s(Z)$, and we note the raw excess and the ratio $T(X)/\bar T(Z)$ alongside $\sigma_T$ wherever conditions are compared. Where a null draws take one value ($s(Z) = 0$) leaves $\sigma_T$ undefined and is reported as such. The standardized excess is a statistic of the pair formed by the cloud and the null ensemble; it changes with the null.

	The decision rule, for a substantive geometric claim about $X$, is that $|\sigma_T(X; Z_{\text{cov}})| \gtrsim 3$, with $Z_{\text{cov}}$ the covariance-matched null, in the direction predicted by the reading of $T$; the threshold is simply a convention: its denominator is the null's spread only, the spread of $T(X)$ across independent subsets of the embedding would add to it.

\subsection{Three nulls}\label{sec:nulls}
	
    The \emph{isotropic null} $Z_{\mathrm{iso}}$ draws $z_i \sim \mathcal{N}(0, I_d)$ for $i = 1, \ldots, n$; it destroys norm scale, anisotropy, all marginals, and all joint structure; an excess against it, by itself, is weak evidence, since anything is registered beyond unstructured noise.

	The \emph{covariance-matched null} $Z_{\mathrm{cov}}$ is the primary null. Let $\hat\mu \in \mathbb{R}^d$ and $\hat\Sigma \in \mathbb{R}^{d \times d}$ be the empirical mean and covariance of $X$; then
	\begin{equation}
		z_i \sim \mathcal{N}(\hat\mu, \hat\Sigma),
	\end{equation}
	implemented through Cholesky $L$ of $\hat\Sigma + 10^{-6}\,\mathrm{tr}(\hat\Sigma)/d \cdot I_d$ with $z_i = L \epsilon_i + \hat\mu$ and $\epsilon_i \sim \mathcal{N}(0, I_d)$. This null preserves the first two moments of $X$, and with them the anisotropy, the principal axes, and the expected squared norm, and destroys cluster structure, multi-modal features, and all higher-order joint statistics. Among all laws on $\mathbb{R}^d$, with mean $\hat\mu$ and covariance $\hat\Sigma$, the Gaussian uniquely maximizes differential entropy relative to Lebesgue measure \citep{cover2006}, so it adds the least to the moments it matches. In (\ref{eq:excess}), it is one-directional. A statistic that exceeds $Z_{\mathrm{cov}}$ detects structure in the data beyond the second moment; a statistic that fails to exceed it is, in these data and at the power of this test, indistinguishable from a functional of the embedding's covariance.

	We must take care, however, in that  when $n \le d$ or $n$ is comparable to $d$, $\hat\Sigma$ is rank deficient, and samples from $\mathcal{N}(\hat\mu, \hat\Sigma)$ are confined to a lower-dimensional subspace than $X$, which biases $\sigma_T$ toward strongly negative values for the dimension estimators and positive values for curvature, since the null appears more concentrated than $X$. For centroid clouds and clusters with $n$ below $2d$ at $d = 300$, we use Ledoit--Wolf shrinkage \citep{ledoit2004},
    
	\begin{equation}
		\hat\Sigma_{\mathrm{LW}} = (1 - \alpha)\hat\Sigma + \alpha \frac{\mathrm{tr}(\hat\Sigma)}{d} I_d,
	\end{equation}
    
	with $\alpha \in [0, 1]$ chosen to minimize an oracle mean-square criterion; $\hat\Sigma_{\mathrm{LW}}$ is full rank for any $n \ge 1$. The shrunk null has a bias of its own: the target is isotropic, so on an anisotropic generator, the null is more isotropic than the data, which moves $\sigma_T$ toward positive values for curvature and negative values for dimension\footnote{See Appendix~\ref{app:calibration} for measures that bias on structureless clouds. }. The point-level null at $n = 2000$, $d = 300$ uses the empirical covariance and no shrinkage; its mismatch, in fact, runs the other way, toward a null more anisotropic than the data, which is conservative for a dimension deficit and for a curvature excess.

	The participation ratio $P = \mathrm{tr}(\hat\Sigma)^2 / \mathrm{tr}(\hat\Sigma^2)$ of a cloud is a functional of its second moments only; a null that misses it by more than $3\sigma$ has failed to reproduce the second moments it was built to reproduce, and its $\sigma_T$ for other statistics inherits a bias of the corresponding sign; a null that matches $P$ can still differ from the data in the rest of the spectrum and in the principal axes. \citet{theiler1996} show that a test drawing typical realizations from a fitted null needs a statistic whose null distribution is the same for every process the null admits. The statistics here depend on $\hat\Sigma$, so the diagnostic bears on every $\sigma_T$.

	The \emph{marginal-permuted null} $Z_{\mathrm{perm}}$ is the diagnostic null; it permutes each coordinate of $X$ independently: for each $j \in \{1, \ldots, d\}$, draw a permutation $\pi_j$ of $\{1, \ldots, n\}$ and set $(Z_{\mathrm{perm}})_{ij} = X_{\pi_j(i), j}$. The univariate marginal distribution of each coordinate is preserved on $\mathbb{R}^d$, and on the sphere up to re-projection; the joint structure, including the covariance, is destroyed. Neither null's preserved properties contain the other's: $Z_{\mathrm{cov}}$ keeps the covariance and makes every marginal Gaussian, and $Z_{\mathrm{perm}}$ keeps every marginal and removes the covariance. A statistic that:
    
    \begin{itemize}
        \item exceeds $Z_{\mathrm{cov}}$ and tracks $Z_{\mathrm{perm}}$ is reproduced by independent coordinates with the data's marginals, and so reflects the marginals;
        \item tracks $Z_{\mathrm{cov}}$ and exceeds $Z_{\mathrm{perm}}$ reflects the covariance;
        \item exceeds both reflects structure maintained by neither null, which can certainly still combine the marginals with the covariance.
    \end{itemize}

\subsection{Frequency-only null}\label{sec:freqnull}

	The above are constructed from the point cloud; training configuration and the corpus are confounded. Let us now consider a fourth null, constructed from the corpus proper, whose tokens have been permuted at random, again, so each word maintains frequency and loses contexts \citep{valentini2022,valentini2023}; the analogous construction for masked language models is the unigram-resampled corpus of \citet{sinha2021}. The resulting embedding preserves the optimizer, the objective, the negative-sampling, the dimensionality, and the unigram distribution, and otherwise destroys the co-occurrence regularities below the scale of the permutation. Valentini and colleagues report that cosine similarity in such embeddings still depends on word frequency, so the null maintains such frequency-driven geometry.

	The share of the original embedding's covariance excess that the shuffled-corpus models also show is  $(T(X_{\mathrm{shuf}}) - \bar T(Z_{\mathrm{cov}}^{\mathrm{shuf}})) / (T(X) - \bar T(Z_{\mathrm{cov}}))$, averaged over shuffle seeds. Under the additivity assumption, a share near one attributes the excess to frequency and configuration, and a share near zero to the corpus. The share is interpretable only where the original's excess passes the decision rule, since a small denominator makes it unstable. The standardized excess of the original against the shuffled ensemble, $\sigma_{\mathrm{shuffle}} = (T(X) - \bar T(X_{\mathrm{shuf}})) / s(T(X_{\mathrm{shuf}}))$, is noted, with the qualification: the denominator comes from three seeds.

\subsection{A note on sphere-normalized variants}\label{sec:sphnull}

	Embeddings used with cosine similarity should be analyzed on the unit sphere, where Euclidean distance is a monotone function of cosine distance. For such embeddings, we $L^2$-normalize rows beforehand, and define $Z_{\mathrm{cov\text{-}sph}}(X_{\mathrm{unit}}) = \mathrm{normalize\_rows} (Z_{\mathrm{cov}}(X_{\mathrm{unit}}))$ matched to the covariance of $X_{\mathrm{unit}}$; draws from $Z_{\mathrm{iso}}$ and $Z_{\mathrm{perm}}$ are re-projected to the sphere in the same way, so that $Z_{\mathrm{iso}}$ becomes the uniform law on the sphere.

	On the sphere, the maximum-entropy law at fixed second moment $\mathbb{E}[x x^\top] = M$ is the Bingham distribution $p(x) \propto \exp(x^\top A x)$ \citep{bingham1974}; row normalization of a Gaussian yields the angular Gaussian, which only approximates it. The discrepancy is, itself, structure beyond the second moment injected by the null, so the maximum-entropy above weakens on the sphere, and the participation-ratio diagnostic checks one summary of the mismatch at the working $(n, d)$. We treat $Z_{\mathrm{cov\text{-}sph}}$ as a convenience, whose adequacy is an assumption.

\section{Estimators and setup}\label{sec:estimators}

	We use a suite summarizing intrinsic dimension, hyperbolicity, curvature, and graph-theoretic properties, to probe for structure in semantic spaces; every statistic is listed with its values against all three nulls in Appendix~\ref{app:full}. Each is applied identically to $X$ and to every null draw.

\subsection{Intrinsic dimension}

	The Levina--Bickel maximum-likelihood estimator \citep{levina2004} is the primary dimension statistic. For each point $x_i$, let $r_k(x_i)$ denote the distance to its $k$-th nearest neighbor. The local estimate at $x_i$ is
	\begin{equation}
		\hat m_k(x_i) = \left[\frac{1}{k-1} \sum_{j=1}^{k-1} \log \frac{r_k(x_i)}{r_j(x_i)}\right]^{-1},
	\end{equation}
	and the global estimate is $\hat d_{\mathrm{MLE}} = \frac{1}{n} \sum_{i=1}^{n} \hat m_k(x_i)$, with $k = 20$. On the covariance-matched nulls of the two released embeddings, with ambient dimension $300$ and $n = 2000$, the estimator reads $61$ to $71$ depending on the spectrum.

	The correlation-dimension estimator \citep{grassberger1983} fits the correlation integral $C(r) = \frac{2}{n(n-1)} \sum_{i < j} \mathbf{1}[\|x_i - x_j\| < r]$, which satisfies $C(r) \sim r^{d_C}$ for small $r$; we fit $\log C(r)$ against $\log r$ at thirty radii spaced logarithmically between the 2nd and 50th percentiles of the pairwise distances and report the slope. Over these radii, $C(r)$ runs from about $0.02$ to $0.5$.

	The graph-based estimator applies the Levina--Bickel formula with $k = 8$ to shortest-path distances on the symmetrized $k$NN graph built at $k = 8$ with Euclidean edge weights, following the intuition that geodesic distance on a manifold is approximated by $k$NN graph distance \citep{tenenbaum2000}; with direct edges dominating the shortest paths to the eight nearest vertices, it is close to a Euclidean estimate at $k = 8$.

	The principal component spectrum gives $\mathrm{dim}_{\mathrm{PCA95}}$, the smallest $K$ such that the top $K$ singular values of the centered $X$ explain at least $95\%$ of the variance, and the participation ratio $P = (\sum_i \lambda_i)^2 / \sum_i \lambda_i^2$ over the eigenvalues $\{\lambda_i\}$ of the empirical covariance. Both are functionals of the second moments; $\mathrm{dim}_{\mathrm{PCA95}}$ is an integer whose null draws coincide at the working $(n, d)$, which leaves its $\sigma$ undefined, and $P$ is the matching diagnostic of Section~\ref{sec:nulls}.

\subsection{Gromov hyperbolicity}

	For four points $(a, b, c, d)$ in a metric space, define the sums $S_1 = d(a,b) + d(c,d)$, $S_2 = d(a,c) + d(b,d)$, $S_3 = d(a,d) + d(b,c)$. The Gromov four-point excess is
    
	\begin{equation}
		\delta(a,b,c,d) = \frac{|S_{(1)} - S_{(2)}|}{2},
	\end{equation}
    
	where $S_{(1)} \ge S_{(2)} \ge S_{(3)}$. A metric space is $\delta$-hyperbolic if $\sup_{a,b,c,d} \delta(a,b,c,d) \le \delta$. For point clouds, we compute the maximum $\delta$ over a uniformly sampled set of $Q$ quadruples and
	
    \begin{equation}
		\delta_{\mathrm{rel}} = \frac{\delta_{\max}^{(Q)}}{\mathrm{diameter}(X)}.
	\end{equation}
    
	This is half the relative hyperbolicity $2\delta/\mathrm{diameter}$ of \citet{khrulkov2020}, which lies in $[0, 1]$. Lower values are associated with more tree-like structure: trees have $\delta = 0$; Euclidean space has $\delta$ proportional to diameter \citep{gromov1987}. We use $Q = 50{,}000$ throughout; the sampled maximum is below the true supremum and stable in $n$ at this $Q$, and the null draws use the same $Q$.

\subsection{Ollivier--Ricci curvature}

	Let $G$ be the symmetrized $k$NN graph of the cloud at $k = 15$, and for each vertex $x$ let $N_k(x)$ be its $k$ nearest neighbors in the directed sense. Define the probability measure $m_x^{\alpha}(y) = \alpha\, \mathbf{1}[y = x] + \frac{1 - \alpha}{k} \mathbf{1}[y \in N_k(x)]$, with $\alpha \in [0, 1]$ the mass left on $x$. The Ollivier--Ricci curvature of an edge $(x, y)$ of $G$ is
    
	\begin{equation}
		\kappa(x, y) = 1 - \frac{W_1(m_x^{\alpha}, m_y^{\alpha})}{d(x, y)},
	\end{equation}
    
	where $W_1$ is the Wasserstein-1 distance \citep{ollivier2007}, and both the transport cost and $d(x, y)$ are ambient Euclidean distances, which on the sphere are chord distances. We use $\alpha = 0.5$, sample $N_{\mathrm{edges}} = 3000$ edges of $G$ uniformly at random, and report mean, standard deviation, and the fraction with $\kappa > 0.1$.

	On a $k$NN graph, $\kappa$ measures the overlap of adjacent neighborhoods. For the combinatorial version of $\kappa$, with hop distances, \citet{jostliu2014} bound it below by an increasing function of the number of triangles on the edge, which on such a graph is the number of shared neighbors; the Euclidean-distance $\kappa$ used here shares that dependence without the proof. A $k$NN graph on a Gaussian cloud already has overlapping neighborhoods, so the null's mean $\bar\kappa$ is positive and an absolute reading of positive $\bar\kappa$ is uninformative by itself. A surviving positive excess indicates adjacent neighborhoods of the data overlap more than those of a Gaussian cloud with the same second moments. 

\subsection{Forman--Ricci curvature}
	Forman--Ricci curvature on an unweighted graph, augmented with the triangle term \citep{forman2003,sreejith2016,samal2018}, is
	\begin{equation}
		F(x, y) = 4 - \deg(x) - \deg(y) + 3 \, |T(x,y)|,
		\label{eq:forman}
	\end{equation}
	where $T(x, y)$ is the set of triangles containing the edge. We use the same symmetrized $k = 15$ graph as for $\kappa$, average over all edges, and report mean and standard deviation of $F$ and the fraction of edges with $F < -10$. Averaged over edges, (\ref{eq:forman}) splits as $\bar F = 4 - \overline{\deg(x) + \deg(y)} + 3\,\overline{|T|}$, with the degree sum averaged over edges (and so weighted toward high-degree vertices); we report both terms alongside $\bar F$ so that an excess in $\bar F$ can be attributed to hubness or to triangle density.

\subsection{Volume growth and ancillary statistics}\label{sec:volume}

	For each of $n_c = 200$ randomly chosen center points $c$, we count $V_c(r) = |\{x \in X : d(c, x) \le r\}|$ at thirty radii spaced logarithmically up to the 30th percentile of the pairwise distances, and average the counts over centers to $\bar V(r)$. The growth exponent is the slope of $\log \bar V(r)$ against $\log r$ over the radii where $1 < \bar V(r) < n/2$; we also fit a log-linear model to detect exponential growth. The exponent over the fitted radii is a dimension estimate at those scales; we consider it alongside the dimension estimators. The growth exponent and the correlation dimension are fit over windows reaching the 30th and the 50th percentile of the pairwise distances, respectively, and the Levina--Bickel and graph-based estimators within $k$NN neighborhoods; we call the first two global and the second two local.

	The maximum degree and Gini coefficient of the degree distribution of the symmetrized $k$NN graph at $k = 15$ complete the suite. High hubness is the canonical mark of distance concentration in high-dimensional Euclidean space \citep{Radovanovi2010}.

\subsection{Embeddings and preprocessing}

	GloVe is the 2024 release of \citet{carlson2025}; it applies the GloVe algorithm \citep{pennington2014glove} to a 220-billion-token subset of Dolma \citep{dolma2024} and has $d = 300$, a vocabulary of $1.2 \times 10^6$, and lowercase tokens. Word2Vec is the pre-trained Google News release \citep{mikolov2013}, with $d = 300$, a vocabulary of $3 \times 10^6$, and cased tokens that include phrases. Both files are frequency-sorted, and we retain the top $10^5$ rows of each.

	On the raw rows of the $n = 2000$ subset, Word2Vec has a norm coefficient of variation of $0.30$ and a Spearman rank correlation between Euclidean and cosine distances over $2 \times 10^4$ sampled pairs of $0.09$: Euclidean and cosine orderings of pairs are nearly unrelated, Euclidean distance is inappropriate, and we sphere-normalize Word2Vec rows before all estimator computations and use $Z_{\mathrm{cov\text{-}sph}}$ for the matched null. GloVe has norm coefficient of variation $0.12$ and rank correlation $0.77$.

\subsection{Sampling and null draws}

	The source file is divided into ten equal-size frequency bands (by row index, which equals the frequency rank for a frequency-sorted source), and equal numbers of rows are drawn from each band. Point-level analyses use $n = 2000$; clustering analyses use $n = 20{,}000$. The two are independent draws from the same $10^5$ rows, and the cluster-removed conditions draw their $n = 2000$ from the remainder of the $20{,}000$ with the same stratification.

	The point-level, cluster-removed, and centroid analyses use $K = 10$ draws per null kind. At the point level and on the cluster-removed subsets, the covariance-matched null is the empirical $Z_{\mathrm{cov\text{-}sph}}$ on the sphere and the empirical $Z_{\mathrm{cov}}$ on the raw rows. At the centroid level, the covariance-matched null is empirical when $n_{\mathrm{cent}} \ge 2d$ and Ledoit--Wolf shrunk otherwise, and the participation-ratio diagnostic is reported with it. Within-cluster statistics are compared with two baselines at the cluster's own sample size: the same statistics on a random subset of the $20{,}000$-point cloud of that size, and, for $k$-means at $k = 100$ and both HDBSCAN settings, a per-cluster covariance-matched null with $K = 5$ draws (empirical when the cluster has at least $2d$ points, shrunk otherwise). On a structureless cloud, $\bar\kappa$ rises and the dimension estimate falls as $n$ shrinks at fixed $k$, which makes these baselines necessary.

\subsection{Corpus and retraining}\label{sec:p3setup}

	The frequency-only null requires retraining, which the released GloVe and Word2Vec models do not permit; contrast is run on a corpus we train on ourselves: the first $5 \times 10^8$ tokens of the English Wikipedia as distributed in the \texttt{wikimedia/wikipedia} dataset (configuration \texttt{20231101.en}), $1{,}004{,}507$ articles in stream order, paragraphs of at least five words as lines, lowercased and tokenized by a word-or-punctuation regular expression: $12{,}193{,}534$ lines and $2{,}585{,}221$ types. Three shuffled versions are produced with independent seeds by permuting the token stream within consecutive blocks of $10^8$ tokens and writing the permuted tokens back into the original line lengths, which keeps every unigram count and every line length; the fraction of adjacent token pairs that recur within the first $5 \times 10^6$ tokens falls from $0.80$ to $0.60$, the residual being the pairs that frequency by itself produces. On the original, and on each shuffled corpus, we train skip-gram with negative sampling with gensim ($d = 300$, window $5$, five negative samples, subsampling threshold $10^{-5}$, minimum count $5$, vocabulary capped at $4 \times 10^5$ types, five epochs, one seed), $391{,}129$ vectors per model, and GloVe with the reference toolkit ($d = 300$, window $15$, $x_{\max} = 100$, $\alpha = 0.75$, learning rate $0.05$, fifteen iterations, minimum count $5$, vocabulary capped at $4 \times 10^5$ types, word and context vectors summed, one seed), eight models in all. The two configurations also differ in window size, subsampling, and number of epochs, and contrasts between them compare configurations. Each model is analyzed with the point-level protocol: top $10^5$ rows, frequency-stratified $n = 2000$, the suite against the three cloud-based nulls with $K = 10$, on the raw rows and on the sphere.

\section{Structure beyond covariance}\label{sec:phase1}

\begin{figure*}
		\centering
		\includegraphics[width=\textwidth]{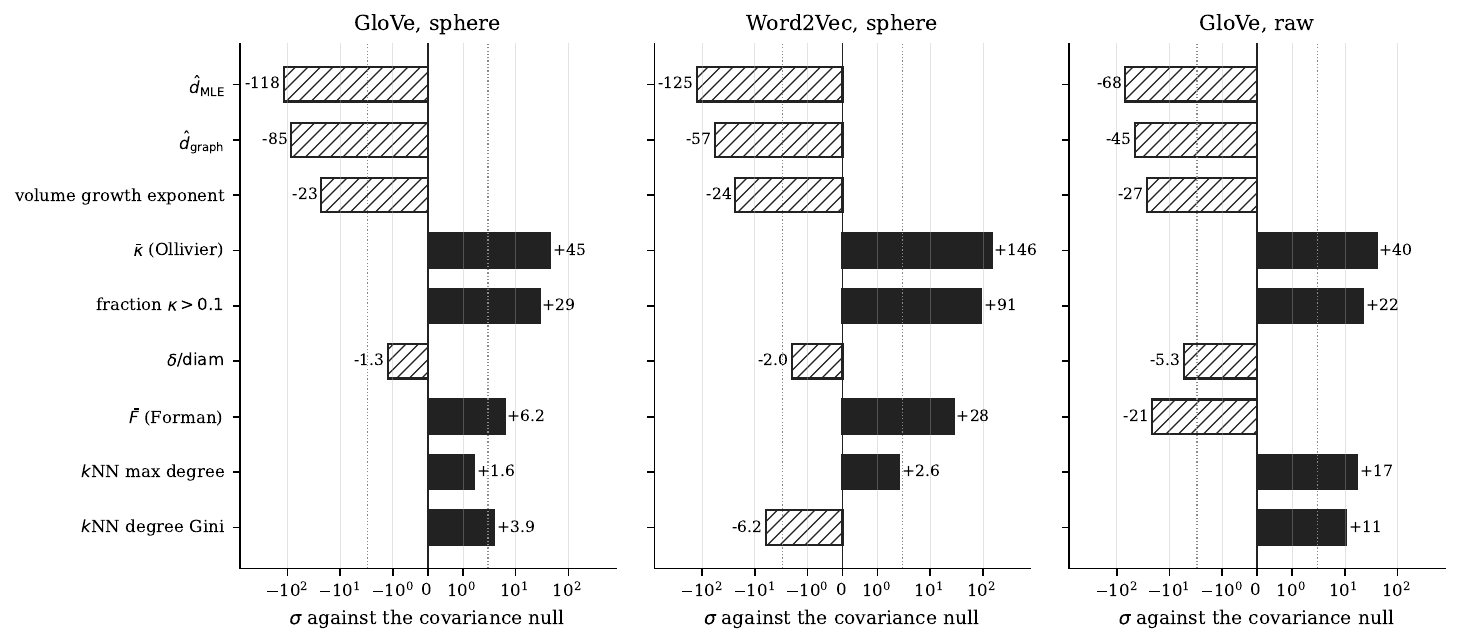}
		\caption{Standardized excess $\sigma$ against the covariance-matched null for nine statistics at $n = 2000$: GloVe on the sphere, Word2Vec on the sphere, and GloVe on the raw rows. Solid bars exceed the null, hatched bars fall below it; the horizontal axis is a symmetric log scale, and dotted lines mark $|\sigma| = 3$.}
		\label{fig:gap_summary}
	\end{figure*}

	At the point level, all statistics are compared with all three cloud-based nulls on $n = 2000$ rows of each embedding. Figure~\ref{fig:gap_summary} plots $\sigma$ against the covariance-matched null for the sphere analyses and for raw GloVe. Table~\ref{tab:phase1_gap} in Appendix~\ref{app:full} gives observed values, null means, and raw excesses for the ten summary statistics, and Table~\ref{tab:full} lists every statistic against every null.

	The dimension estimators and the Ollivier summaries exceed the null in the predicted direction: the Levina--Bickel estimate is $\hat d_{\mathrm{MLE}} = 25.9$ (GloVe) and $27.2$ (Word2Vec), against null values of $60.6$ and $71.3$; $\sigma$ is $-118$ and $-125$. The graph-based and correlation estimators and the volume growth exponent give the same direction at $|\sigma|$ between $23$ and $85$. The absolute reading $\hat d \approx 26$ is a value on the estimator's compressed scale, on which the full-rank covariance-matched nulls read $61$ to $71$ at this $n$ and $k$; the table establishes the contrast against the null and the direction of every dimension estimator. Ollivier $\bar\kappa$ is $0.167$ and $0.134$ against null means of $0.075$ and $0.052$, ratios $2.2$ and $2.6$, $\sigma = +45$ and $+146$; the fraction of edges with $\kappa > 0.1$ moves the same way. The positive null means are the graph-level baseline of Section~\ref{sec:estimators}, the overlap that a $k$NN graph produces on a Gaussian cloud of this covariance.

	Forman $\bar F$ is $-51.7$ (GloVe) against a null of $-58.8$, $\sigma = +6.2$, and $-25.9$ (Word2Vec) against $-38.6$, $\sigma = +28$: the embeddings have less negative Forman curvature than the null, so a strongly negative $\bar F$ as embedding-specific hub-and-bridge structure is not the case. The decomposition of (\ref{eq:forman}) is the excess (see Appendix~\ref{app:full}). In GloVe, the edge-averaged degree sum is $77.7$ against $70.1$ in the null, a term that lowers $\bar F$ by $7.6$, while the triangle term is $7.34$ against $2.43$, three times the null, and raises $\bar F$ by $14.7$; the net excess of $+7.1$ is the triangle term less the degree term. In Word2Vec, the degree sum is $44.0$ against $45.7$ (a contribution of $+1.7$) and the triangle term $4.72$ against $1.05$, four and a half times the null (a contribution of $+11.0$). The Forman excess is, therefore, triangle density, the same local clustering that Ollivier $\kappa$ measures, seen through the other discretization, the relation \citet{samal2018} report between the two curvatures on complex networks. In GloVe, the degree term runs the way a hub-and-bridge reading predicts and is outweighed; in Word2Vec, it falls below the null ($-3.1\sigma$), against that reading.

	The $k$NN degree Gini exceeds the null with opposite signs, $+3.9$ on GloVe and $-6.2$ on Word2Vec; the maximum degree ($257$ against $195$ and $113$ against $83$; $\sigma = +1.6$ and $+2.6$) and $\delta/\mathrm{diameter}$ ($0.095$ against $0.107$ and $0.086$ against $0.100$; $\sigma = -1.3$ and $-2.0$) track the null. Reading low absolute $\delta/\mathrm{diameter}$ as ``tree-likeness'', which was an initial hypothesis of the project, requires more than this statistic, however, since a covariance-matched sphere null produces a value within $2\sigma$ of the observed one; $k$NN hubness as embedding-specific structure is likewise, consistent with hubness as a property of high-dimensional Euclidean and angular spaces \citep{Radovanovi2010}. Two statistics leave $\sigma$ undefined or otherwise unstable: $\mathrm{dim}_{\mathrm{PCA95}}$ takes one value on every null draw ($216$ and $245$), and the fraction of edges with $F < -10$ is $1.000$ on the Word2Vec draws and within $10^{-5}$ of it on the GloVe draws. For the second, the observed fractions of $0.933$ and $0.915$ fall below the null's, in the direction of the Forman result. The participation ratio, the second-moment diagnostic, is $+1.8\sigma$ for GloVe and $+13.0\sigma$ for Word2Vec, the null more anisotropic than the data in both cases; the Word2Vec value fails the $3\sigma$ criterion of Section~\ref{sec:nulls} in the conservative direction.

	Against the isotropic null, every statistic exceeds the threshold on GloVe, and every one except $\delta/\mathrm{diameter}$ ($+0.7\sigma$) on Word2Vec, which is the weak evidence in Section~\ref{sec:nulls}. Against the marginal-permuted null, two of the ten summary statistics, both of them degree statistics, diverge from the covariance-matched comparison (see Appendix~\ref{app:full}). Word2Vec's maximum degree tracks $Z_{\mathrm{cov\text{-}sph}}$ at $+2.6\sigma$ and exceeds $Z_{\mathrm{perm}}$ at $+8.5\sigma$, the pattern of a statistic that reflects the covariance (see Section~\ref{sec:nulls}). GloVe's degree Gini reads $+3.9\sigma$ against $Z_{\mathrm{cov\text{-}sph}}$ and $-19\sigma$ against $Z_{\mathrm{perm}}$. Hubness depends on the anisotropy otherwise destroyed by the permuted null.

\subsection{Remarks on raw-row analysis of GloVe}\label{sec:raw}
    
	$\delta/\mathrm{diameter}$ is $0.047$ against a null of $0.097$, $\sigma = -5.3$; $\bar F$ is $-250$ against $-85$, $\sigma = -21$; the maximum degree is $1553$ against $386$, $\sigma = +17$; the Gini is $0.378$ against $0.313$, $\sigma = +11$. Each exceeds the covariance-matched null: a star-like $k$NN graph around a few hubs is tree-like in $\delta$, hub-dominated in $F$, and hub-dominated in degree. The Forman decomposition shows that the degree sum is $283$ against $97$ in the null, and the triangle term $9.8$ against $2.6$, so both terms are elevated and the degree term dominates, the reverse of the sphere! The dimension and Ollivier findings persist on the raw rows ($\sigma = -68$ for $\hat d_{\mathrm{MLE}}$, $+40$ for $\bar\kappa$, ratio $2.8$). 
    
    The projection to the sphere discards the norm distribution (GloVe norms vary with coefficient of variation $0.12$ on the raw rows) and with it the structure behind the reversal. High-dimensional hubness concentrates on points near the cloud's mean \citep{Radovanovi2010}, so the raw-row hubs are (plausibly) the rows closest to the mean vector; the identity of the hub words remains unverified, however, and available for future work.

\section{Attribution by retraining}\label{sec:phase3}
    
\begin{figure*}
		\centering
		\includegraphics[width=\textwidth]{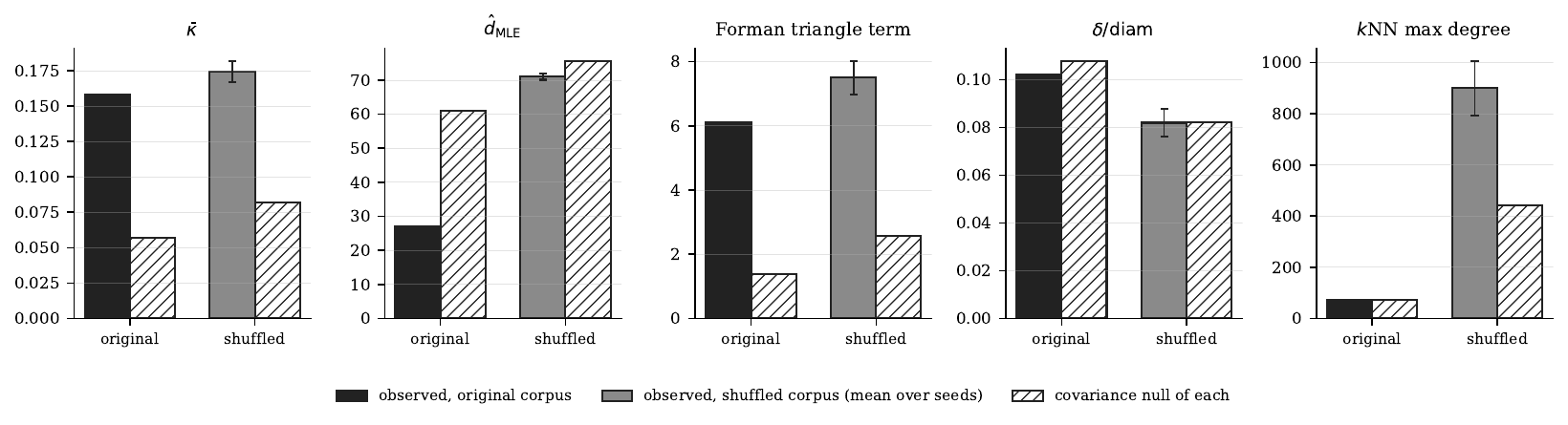}

		\medskip
		\includegraphics[width=\textwidth]{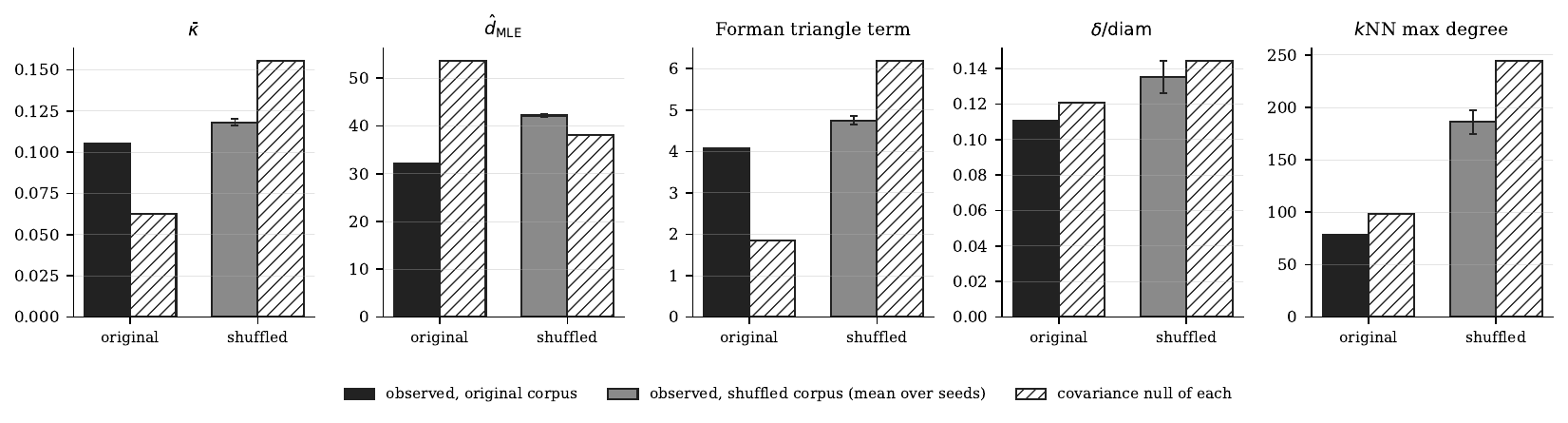}
		\caption{Skip-gram (top) and GloVe (bottom) on the original and shuffled corpora over observed values (solid; shuffled-corpus models averaged over seeds with their spread) beside covariance-matched null of each (hatched).}
		\label{fig:shuffled}
	\end{figure*}

	Table~\ref{tab:shuffled_selected} and Figure~\ref{fig:shuffled} show the contrast of Section~\ref{sec:freqnull} for skip-gram and GloVe on the Wikipedia slice, on the sphere; Appendix~\ref{app:retraining} the full tables on the sphere and on the raw rows. The contrast attributes the local dimension deficit to co-occurrence for both configurations; it leaves undertermined the source of skip-gram's curvature excess, because the shuffled-corpus skip-gram models reach an excess of the same size through a different graph.

	\begin{table*}
		\centering
		\small
		\setlength{\tabcolsep}{4pt}
		\begin{tabular}{lrrrrrr}
        \toprule
         & \multicolumn{3}{c}{\textbf{Skip-gram}} & \multicolumn{3}{c}{\textbf{GloVe}} \\
        \cmidrule(lr){2-4} \cmidrule(lr){5-7}
        \textbf{Statistic} & \textbf{Original excess} & \textbf{Shuffled excess} & \textbf{Share} & \textbf{Original excess} & \textbf{Shuffled excess} & \textbf{Share} \\
        \midrule
        \tabhead{$\hat d_{\mathrm{MLE}}$ (local)} & $-33.8$ & $-4.56$ & $0.13$ & $-21.6$ & $4.03$ & $-0.19$ \\
        \tabhead{$\hat d_{\mathrm{graph}}$ (local)} & $-41.6$ & $-3.85$ & $0.09$ & $-29.0$ & $4.40$ & $-0.15$ \\
        \tabhead{$\hat d_{\mathrm{corr}}$ (global)} & $-7.26$ & $-9.35$ & $1.29$ & $-3.73$ & $-4.42$ & $1.19$ \\
        \tabhead{Volume growth exponent (global)} & $-14.2$ & $-12.5$ & $0.88$ & $-7.30$ & $-6.31$ & $0.87$ \\
        \tabhead{$\bar\kappa$ (Ollivier)} & $0.101$ & $0.0924$ & $0.91$ & $0.0427$ & $-0.0373$ & $-0.87$ \\
        \bottomrule
        \end{tabular}
		\caption{Excess over the covariance-matched null for the original model and for the shuffled-corpus models (mean over three seeds, each against its own null), and the share of the original's excess shared by the shuffled-corpus models.}
		\label{tab:shuffled_selected}
	\end{table*}

	Both Wikipedia-trained models reproduce the point-level picture of the released models on the sphere: skip-gram reads $\hat d_{\mathrm{MLE}} = 27.2$ against a covariance-matched null of $61.0$ and $\bar\kappa$ at $2.8$ times its null, and GloVe at this corpus size reads $32.1$ against $53.7$ and $1.7$ times its null. Both show the same triangle excess in $\bar F$ and the same tracking of the null by $\delta$ and hubness as in Section~\ref{sec:phase1}; on the raw rows both show the raw-GloVe pattern of Section~\ref{sec:raw}. Results from the shuffled-corpus models, therefore, bear on the released ones through this correspondence.

	The two configurations produce opposite clouds from a context-free corpus. Skip-gram's shuffled-corpus models are nearly isotropic (participation ratio $256$ of $300$, Ledoit--Wolf weight $0.96$), with hub-dominated $k$NN graphs. GloVe's shuffled-corpus models collapse onto a few directions: participation ratio $11$ on the sphere and $2.3$ on the raw rows, correlation dimension $6$, norm coefficient of variation $0.70$. With every context distribution equal to the unigram distribution, the log co-occurrence that GloVe factorizes is a function of the two frequencies only. The low participation ratio and correlation dimension suggest that the vectors lie along a frequency curve, which a plot of the leading coordinates against log frequency would confirm. Each shuffled model is compared with its own covariance-matched null, and for GloVe that null is a thin Gaussian whose $k$NN graph is itself dense ($\bar\kappa = 0.156$ on the null against $0.05$ to $0.08$ for every other sphere null at $n = 2000$), so GloVe's shares measure an excess over a reference unlike those of the originals.

	The shuffled-corpus models of both configurations lose the local dimension deficit: skip-gram's read $\hat d_{\mathrm{MLE}} = 71.1 \pm 0.9$ against their null of $75.6$, an excess of $-4.6$ where the original has $-33.8$: $13\%$ retained, the original $46$ shuffled standard deviations below them, with the graph estimator at $9\%$. GloVe's $42.2 \pm 0.2$ against $38.2$, above their null, where the original reads $32.1$ against $53.7$: a share of $-0.19$, the original $41$ shuffled standard deviations below. On a context-free corpus, both configurations produce clouds whose local dimension lies within $5$ of their null's, against deficits of $21.6$ to $33.8$ in the intact-corpus models. On this corpus, co-occurrence accounts for most of the deficit behind the readings of $27.2$ and $32.1$. The two global dimension statistics behave differently in both configurations: the correlation dimension deficit is reproduced ($9.4$ against $7.3$ for skip-gram; $4.4$ against $3.7$ for GloVe) and so is $87$ to $88\%$ of the volume growth deficit. Both are fit over windows that extend far beyond the $k$NN scale, and both respond to the concentration of a shuffled cloud around a common direction or along a curve, which frequency by itself produces. The dimension family of Section~\ref{sec:phase1} thus splits: the two local estimators track co-occurrence, and the two global ones track frequency and configuration.

	Skip-gram's shuffled-corpus models reproduce the Ollivier curvature excess, and GloVe's fall below their null: skip-gram's read $\bar\kappa = 0.175 \pm 0.007$ against their null of $0.082$, an excess of $0.092$ where the original has $0.101$: $91\%$ of the excess, the original inside the shuffled ensemble ($\sigma_{\mathrm{shuffle}} = -2.2$), the fraction of edges with $\kappa > 0.1$ higher than the original's ($0.96$ against $0.65$), and the Forman triangle term at $104\%$ of its excess. GloVe's read $\bar\kappa = 0.118 \pm 0.002$ against a null of $0.156$, below it, where the original has $0.105$ against $0.063$. In absolute terms the original lies $6.5$ shuffled standard deviations below the shuffled models. The two excesses differ because their nulls differ. An embedding trained on a context-free corpus can, therefore, show the curvature excess over the covariance-matched null at the original's magnitude, as so does the shuffled-corpus skip-gram. For skip-gram, the size of the excess, thus, leaves its source undetermined. For GloVe, the shuffled-corpus models sit below their own null, and the original's excess passes this test as co-occurrence structure. GloVe's context-free geometry differs so far from the original's that the two excesses are measured against different references, and the GloVe attribution is a bound. The shuffled-corpus skip-gram models reach the excess through a different graph, dominated by hubs: maximum degree $900 \pm 106$ against the original's $74$, degree Gini $0.42$ against $0.18$, an edge-averaged degree sum of $296$ against $50$, and $\bar F = -269$ against $-27.5$; their neighborhoods overlap around a few vertices near the common direction, and Ollivier $\kappa$ at $\alpha = 0.5$ counts that overlap as curvature. The original reaches the same excess with modest hubness and a local dimension of $27$, so its overlap is spread across the cloud. The joint pattern of low local dimension and rather modest hubness appears only in the models trained on the intact corpus.

	On the raw rows, the shuffled-corpus models of both configurations reproduce the tree-like $\delta$ of Section~\ref{sec:raw}, at $122\%$ of the excess for skip-gram and $110\%$ for GloVe, so frequency and norm by themselves produce it. Shuffled-corpus skip-gram also produces the raw-row hubs and the hub-and-bridge $\bar F$, which in GloVe need co-occurrence; Appendix~\ref{app:retraining} gives the raw-row tables.

	Some disclaimers follow. The contrast was run on one corpus with one original model per configuration and three shuffles; the released models inherit the attribution through the correspondence above, without retraining. Skip-gram's shuffled-corpus models reach the original's curvature with a hubness otherwise lacking in the original, and GloVe's reach a geometry never even approached by the original. For the purely combinatorial Forman terms, degree-preserving rewiring of the $k$NN graph supplies part of such a null. The shuffled-corpus models' own covariance-matched nulls are mismatched by the participation-ratio diagnostic ($+66$ to $+95\sigma$ for skip-gram in the conservative direction, so its $\bar\kappa$ excess is if anything understated; $-6.6\sigma$ for GloVe, whose shuffled-corpus models are thinner than their null).

\section{Robustness across clusters and scales}\label{sec:phase2a}

	The local dimension deficit and the Ollivier curvature excess could well come from discrete clusters, whose local geometry are averaged over by the global statistics, or from properties of the whole embedding itself. A mixture finer than the clusterings tested, a stratified space, and clusters beside a diffuse component also fit the global statistics.

\subsection{Clustering}

	We run Euclidean $k$-means on the sphere-normalized stratified subset of $n = 20{,}000$ points at $k \in \{50, 100, 200, 400, 800, 1000\}$, with ten initializations and a fixed random state, and renormalize the centroids to the unit sphere. This approximates spherical $k$-means, and assignments can differ at cluster boundaries because Euclidean assignment to an unnormalized centroid depends on its norm. HDBSCAN \citep{campello2013,mcinnes2017} runs on the same subset at \texttt{min\_cluster\_size} $\in \{50, 100\}$, with \texttt{min\_samples} tied to it as in the library default. The centroids are the normalized means of the clusters' members, with noise points excluded. The centroid tests maintain clusters of at least $50$ members at $k \le 200$ and at least $10$ at $k \ge 400$. The centroid count $n_{\mathrm{cent}}$ then reaches $2d$ at $k \ge 800$, where the empirical covariance-matched null applies.

\subsection{Within-cluster geometry}\label{sec:within}

	For each clustering with mean cluster size at least $100$, we compute the suite on the twenty largest clusters. At fixed $k$ the estimators move with $n$, so these values need a baseline at their own sample size. On a structureless anisotropic Gaussian cloud projected to the sphere, with a spectrum close to GloVe's, $\bar\kappa$ rises from $0.067$ at $n = 2000$ to $0.231$ at $n = 50$, and $\hat d_{\mathrm{MLE}}$ falls from $63.8$ to $31.7$ (see Appendix~\ref{app:clusters}).

\begin{figure*}
		\centering
		\includegraphics[width=\textwidth]{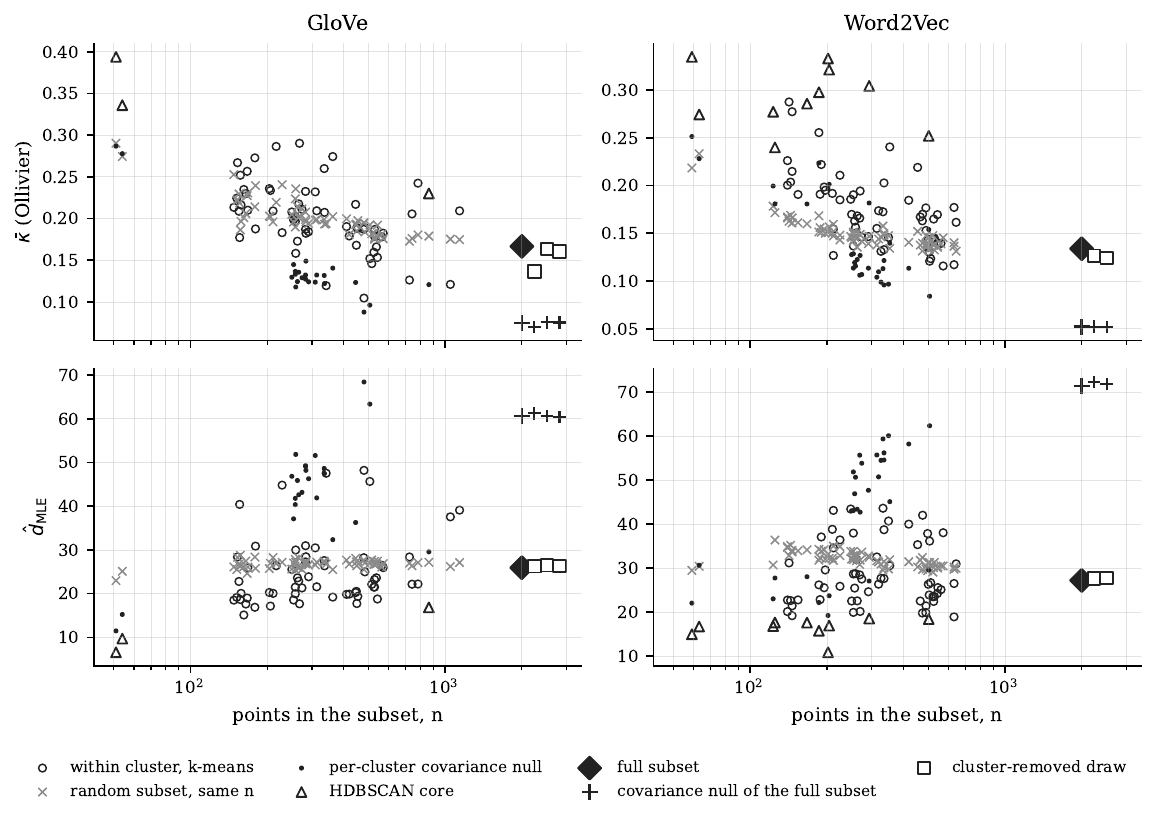}
		\caption{$\bar\kappa$ (top) and $\hat d_{\mathrm{MLE}}$ (bottom) against subset size $n$ for every subsetting condition, per embedding. Open circles: $k$-means clusters at $k \in \{50, 100, 200\}$; triangles: HDBSCAN clusters; crosses: random subsets of the cloud at the same sizes; dots: per-cluster covariance-matched nulls, shrunk below $2d$ points; diamond: the full $n = 2000$ subset with its null as a plus sign; squares: the cluster-removed draws with their nulls, offset in $n$ for legibility.}
		\label{fig:stat_vs_n}
	\end{figure*}

	Each cluster is compared with a random subset of the $20{,}000$-point cloud at its own size and, at $k = 100$ and for HDBSCAN, with a per-cluster covariance-matched null. The per-cluster nulls are shrunk below $2d$ points and biased toward a curvature excess (see Appendix~\ref{app:calibration}), so the comparison rests on the random subsets. Figure~\ref{fig:stat_vs_n} shows every cluster against both baselines as a function of $n$, and Table~\ref{tab:within} in Appendix~\ref{app:clusters} gives medians across clusters.

	Against the random subsets, $k$-means clusters in GloVe are as curved as the cloud around them: at $k = 50$, $100$, and $200$ their median $\bar\kappa$ lies within about $5\%$ of the controls', while their median $\hat d_{\mathrm{MLE}}$ lies $8$ to $25\%$ below. In Word2Vec, the clusters exceed their controls in $\bar\kappa$ by $8$, $12$, and $27\%$ at the same $k$, and their $\hat d_{\mathrm{MLE}}$ lies $11$ to $18\%$ below. A $k$-means cell is therefore a region of lower local dimension than a random subset of its size in both embeddings; its curvature matches the cloud's in GloVe, and then exceeds it by a fraction of the global excess in Word2Vec. Against the $n = 2000$ value, GloVe's clusters at $k = 100$ read $1.23$ times as curved, a difference that the sample-size effect accounts for.

	The HDBSCAN clusters are more curved than random subsets of their size: in Word2Vec the clusters of $186$ to $502$ points have a median $\bar\kappa$ of $0.298$ against a control of $0.151$, and a median $\hat d_{\mathrm{MLE}}$ of $17.0$ against $32.2$. In GloVe the $863$-point high-frequency cluster reads $\bar\kappa = 0.230$ against a control near $0.18$, and $\hat d_{\mathrm{MLE}} = 16.9$ against $27$. The two numeric GloVe clusters, of $51$ and $54$ points, lie below the regime of the $k = 20$ dimension estimator.

\subsection{Cluster contents}

	Table~\ref{tab:cluster_types} summarizes the HDBSCAN clusters at \texttt{min\_cluster\_size} $= 50$, with our characterizations of the frequency-ordered word lists. At $100$, every GloVe point is labeled noise, and Word2Vec re-extracts three clusters of $502$, $204$, and $186$ points, the clusters behind the Word2Vec column of Table~\ref{tab:within}. Cluster counts depend on the density and extraction settings of HDBSCAN (see Appendix~\ref{app:clusters}).

	\begin{table}
		\centering
		\footnotesize
		\begin{tabular}{>{\raggedright\arraybackslash}p{3.3cm}rr>{\raggedright\arraybackslash}p{9.2cm}}
        \toprule
        \textbf{Model} & \textbf{Clusters} & \textbf{Points} & \textbf{Cluster contents} \\
        \midrule
        \tabhead{GloVe} & $3$ & $968$ & high-frequency vocabulary; small decimal numbers; larger decimal numbers \\
        \tabhead{Word2Vec} & $7$ & $1{,}032$ & news-wire boilerplate; first names; surnames; dateline cities; NBA and NHL teams and players; MLB players; football, cricket, and South Asian place names \\
        \tabhead{Wikipedia skip-gram} & $5$ & $808$ & years; three-digit numbers; Japanese names; Indian personal and place names; biochemical vocabulary \\
        \tabhead{Wikipedia GloVe} & $3$ & $917$ & function words with numerals; place and personal names; evaluative adjectives and abstract nouns \\
        \tabhead{Shuffled-corpus models, seed 1} & $0$ & $0$ & every point labeled noise \\
        \bottomrule
        \end{tabular}
		\caption{HDBSCAN clusters at \texttt{min\_cluster\_size} $= 50$ on the $20{,}000$-point subsets, with our characterizations of the frequency-ordered word lists.}
		\label{tab:cluster_types}
	\end{table}

	We call a cluster a format list when its members share a document format, as datelines or box scores do, and a taxonomic class when they share a category, as colors or animals do. note that the descriptions may well overlap, since names and cities form taxonomic classes and also fill attribution lists and datelines. Colors, animals, emotions, body parts, and kinship terms are absent from the clusters of both released models. The listed words suggest shared training-corpus contexts: athletes listed together in box scores, boilerplate beside other boilerplate, names from attribution lists, place names from dateline formats, and, in GloVe, high-frequency vocabulary that neighbors itself by ubiquity.

	In the Wikipedia-trained models, two of eight clusters are as topical classes: biochemical vocabulary in skip-gram, and evaluative adjectives and abstract nouns in GloVe; the format reading fits both poorly. The high-frequency cluster of GloVe on Dolma recurs in GloVe on Wikipedia and is absent from skip-gram on the same corpus, so it belongs to the GloVe configuration. On both shuffled-corpus models, HDBSCAN labels every point as noise at this setting, in skip-gram's case beside twelve times the original's maximum degree. At the settings tested here, the clusters, therefore, depend on co-occurrence. Dense clusters and hubs are, thus, separate phenomena.

\subsection{Cluster-removed subsets}\label{sec:diffuse}

	If the clusters have curvature signal, then removing them should collapse the point-level excess. We remove the clusters from the $20{,}000$-point subset, draw $n = 2000$ points from the remainder with the frequency stratification of the point-level subset, and rerun the three-null analysis. The conditions remove all clusters at \texttt{min\_cluster\_size} $= 50$ and at $100$, and, for GloVe, the two numeric clusters only. At $100$ HDBSCAN labels every GloVe point as noise, so that condition is a replicate draw of the full one. Because the null's standard deviation varies across conditions, Figure~\ref{fig:diffuse_preservation} and Table~\ref{tab:diffuse} in Appendix~\ref{app:clusters} show the raw excess and the ratio beside $\sigma$.

	\begin{figure*}
		\centering
		\includegraphics[width=\textwidth]{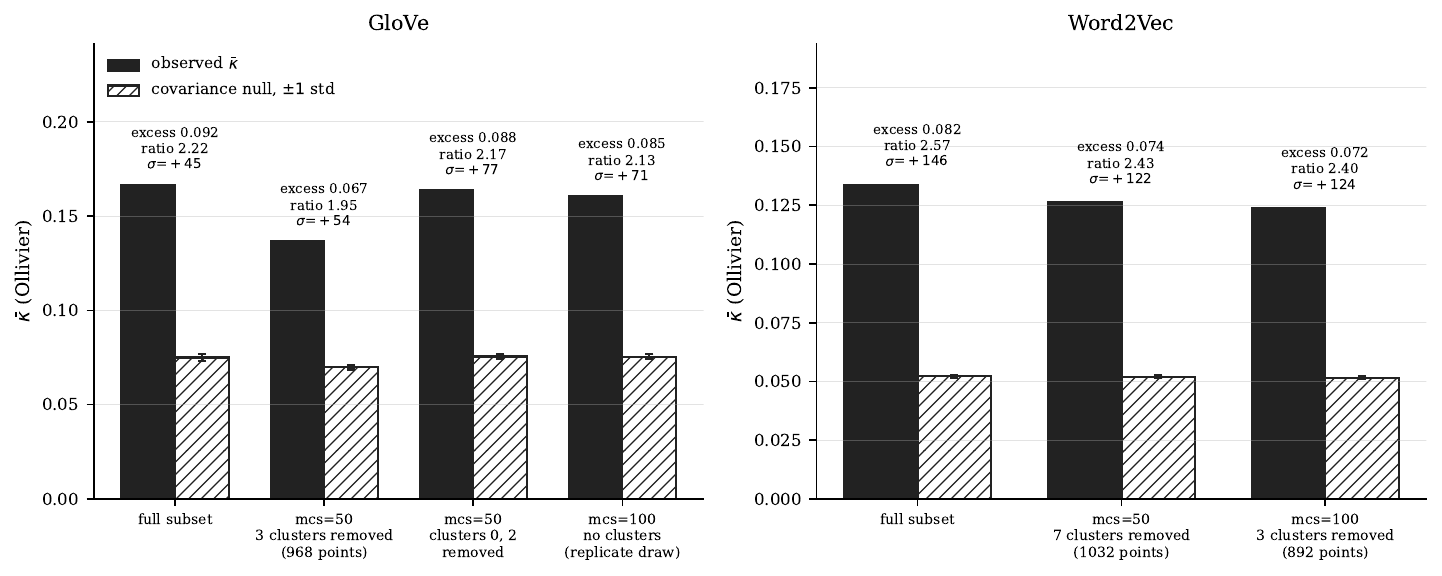}
		\caption{Ollivier $\bar\kappa$ across the conditions of Table~\ref{tab:diffuse}. Solid bars: observed $\bar\kappa$; hatched bars: covariance-null mean with $\pm 1$ standard deviation over the $K$ draws. Annotations are excess, ratio, and $\sigma$.}
		\label{fig:diffuse_preservation}
	\end{figure*}

	Removing the three GloVe clusters leaves $73\%$ of the raw excess, $0.067$ against $0.092$ on the full subset, and removing only the two numeric clusters leaves $96\%$. The reduction, thus, comes from the high-frequency cluster, whose $4.3\%$ of the points hold about a quarter of the raw excess. In Word2Vec, removing the seven clusters at \texttt{min\_cluster\_size} $= 50$ leaves $91\%$ of the excess, and removing the three at $100$ leaves $89\%$. After removal, $\sigma$ rises in GloVe, to $+54$ and $+77$ against $+45$, because the null's standard deviation is lower on these draws than on the full subset. The replicate draw of the full condition reads $+71$. The local dimension deficit and the triangle excess of $\bar F$ persist in every condition: $\hat d_{\mathrm{MLE}}$ stays within $0.6$ of its full-subset value, and the triangle term stays at $2.6$ to $4.3$ times its null.

	The GloVe replicate draw $\bar\kappa = 0.161$, an excess of $0.085$, against $0.167$ and $0.092$. Two draws of one condition thus differ by $0.006$ in $\bar\kappa$, three times the full subset's null standard deviation. Differences in $\sigma$ of that order between conditions are therefore draw noise. The GloVe reduction, after removing all three clusters, $0.025$ in excess, is four times that difference, and the numeric-only and all-cluster conditions differ by $0.021$, in the same direction, on independent draws. The Word2Vec reductions are $1.1$ and $1.4$ by it, so at this replication Word2Vec retention is indistinguishable from full retention.

	Removing GloVe's three clusters also lowers its hubness: the maximum degree drops from $257$ to $105$, below its null, and the degree Gini moves from $+3.9\sigma$ to $-5.5\sigma$. The numeric-only condition keeps both. The $k$NN hubs of the GloVe subset thus sit in the high-frequency cluster. Across the GloVe conditions $\delta/\mathrm{diameter}$ reads $\sigma$ between $-4.4$ and $-0.4$, and in Word2Vec it reads $-2.4$ and $-2.2$. Two draws of one condition differ by $3\sigma$ in this statistic, so draw noise covers that spread. Every value is negative, as at the point level.

\subsection{Centroid test}\label{sec:centroids}

	If the embedding has hierarchical structure above the cluster level, then the cluster centroids should, themselves, be approximately tree-like: the global Gromov $\delta / \mathrm{diameter}$ on the centroid point cloud should be low relative to a covariance-matched null fit to the centroids.

	\begin{table*}
		\centering
		\small
		\setlength{\tabcolsep}{4pt}
		\begin{tabular}{lrrrrrrrr}
        \toprule
        \textbf{Clustering} & $\boldsymbol{n_{\mathrm{cent}}}$ & $\boldsymbol{\delta_{\mathrm{rel}}}$ & $\boldsymbol{\sigma_{\delta}}$ & $\boldsymbol{\bar\kappa}$ & $\boldsymbol{\bar\kappa}$ \textbf{null} & $\boldsymbol{\sigma_{\bar\kappa}}$ & $\boldsymbol{\sigma_{\hat d}}$ & $\boldsymbol{\sigma_{P}}$ \\
        \midrule
        \tabhead{GloVe $k$-means $k=800$} & 737 & $0.127$ & $-0.8$ & $0.178$ & $0.102$ & $+69$ & $-42$ & $+2.2$ \\
        \tabhead{GloVe $k$-means $k=1000$} & 851 & $0.108$ & $-1.8$ & $0.175$ & $0.101$ & $+40$ & $-46$ & $+1.9$ \\
        \tabhead{Word2Vec $k$-means $k=800$} & 771 & $0.133$ & $+1.4$ & $0.164$ & $0.0870$ & $+48$ & $-72$ & $+5.1$ \\
        \tabhead{Word2Vec $k$-means $k=1000$} & 909 & $0.119$ & $-0.4$ & $0.161$ & $0.0826$ & $+53$ & $-60$ & $+5.1$ \\
        \bottomrule
        \end{tabular}
		\caption{Centroid-level tests at $k \ge 800$, where $n_{\mathrm{cent}} \ge 2d$; $n_{\mathrm{cent}}$ counts centroids of clusters with at least $10$ members. Observed $\delta/\mathrm{diameter}$ and $\bar\kappa$ with the null mean of $\bar\kappa$; $\sigma_{\delta}$,$\sigma_{\bar\kappa}$, and $\sigma_{\hat d}$ are the standardized excesses of $\delta/\mathrm{diameter}$, $\bar\kappa$, and $\hat d_{\mathrm{MLE}}$, and $\sigma_{P}$ that of the participation ratio $P$, the second-moment diagnostic of Section~\ref{sec:nulls}.}
		\label{tab:centroid_tree}
	\end{table*}

	Table~\ref{tab:centroid_tree} gives the four clusterings, whose centroid count reaches $2d$ (empirical null applies), and Appendix~\ref{app:calibration} the rows with shrunk nulls. On the four rows $\sigma_\delta$ lies between $-1.8$ and $+1.4$, inside the threshold. The rows share data and count as fewer than four independent tests, so the per-row tests leave the hierarchy hypothesis undecided at this scale, left for future work.

	At $k = 800$ and $1000$ the participation-ratio diagnostic reads $+1.9$ to $+5.1$, inside the threshold or on its conservative side. Against that null the centroid $\bar\kappa$ is $1.7$ to $1.9$ times the null mean, at $\sigma$ between $+40$ and $+69$, and $\hat d_{\mathrm{MLE}}$ is about $0.3$ times its null's, at $\sigma$ between $-42$ and $-72$. The centroid cloud, thus, shows the point cloud's two signals at comparable strength, against point-level ratios of $2.2$ to $2.6$ for $\bar\kappa$ and $0.4$ for $\hat d_{\mathrm{MLE}}$.

\section{Discussion}\label{sec:discussion}

	The results here are on-going on the inference from a statistic's value to structure beyond covariance, and on the further inference from that structure to its source in the corpus.

\subsection{Detection and attribution}

	The statistics that concern covariance-matched null, $\delta/\mathrm{diameter}$ and the maximum degree, agree in these data with what is produced by the covariance. An absolute $\delta/\mathrm{diameter}$ of $0.086$ to $0.095$, or $0.17$ to $0.19$ on the scale of \citet{khrulkov2020}, lies close to the null's value at the same $(n, d)$. For the released embeddings, its sign is negative at the point level and after cluster removal, and varies across the centroid clouds. High $k$NN hubness is consequence of distance concentration in high-dimensional spaces \citep{Radovanovi2010}.

	We interpret curvature excess as overlap of adjacent neighborhoods, beyond what is produced by a Gaussian cloud with the embedding's covariance. The overlap extends across the cloud, at a local dimension below the Gaussian's. As Ricci curvature awaits an estimator with a length scale, not developed here. On a context-free corpus, the skip-gram configuration reaches the same overlap through hubs, while on the intact corpus, co-occurrence spreads it across a low-dimensional cloud.

	Against the covariance-matched null, $\bar\kappa$ exceeds at $+45\sigma$ and $+146\sigma$; against a null that also preserves the training configuration and the unigram distribution, the original skip-gram value sits inside the shuffled ensemble ($\sigma_{\mathrm{shuffle}} = -2.2$), and the original GloVe value sits below it ($\sigma_{\mathrm{shuffle}} = -6.5$). GloVe's excess over its own null survives the contrast. The test credits that excess to the corpus only as a bound, since the shuffled models are measured against a thinner null than the original's. A Gaussian null supports a claim of structure beyond the second moments; accounting for the source of that structure, however, needs a null that keeps the generating process and removes the signal of interest, and for trained representations that null comes from retraining on a corpus with the signal destroyed. A statistic that two different graphs can satisfy, as $\bar\kappa$ is satisfied by overlap around hubs and by overlap spread across the cloud, must be considered with the statistics that distinguish those graphs (here the degree terms and the local dimension).

\subsection{Dependence on metric and scale}

	On the raw GloVe rows, $\delta/\mathrm{diameter}$, Forman curvature, and both degree statistics differ from the null. On the raw rows of the Wikipedia models, the shuffled-corpus models of both configurations reproduce the low $\delta/\mathrm{diameter}$, so frequency and norm suffice to produce the value that a hierarchical reading would use. Attribution depends on scale within the dimension family: the local estimators, Levina--Bickel and graph-based, track co-occurrence, and the global ones, the correlation dimension and the volume growth exponent, track frequency and configuration.

	The estimators also move with sample size at fixed $k$: on a structureless cloud, $\bar\kappa$ rises, and the dimension estimate decreases as $n$ decreases. Every within-cluster value is, therefore, compared with a random subset of its own size. The two statistics move together on structureless data, and are computed on nested $k$NN neighborhoods ($k = 15$ for $\bar\kappa$ and $k = 20$ for $\hat d_{\mathrm{MLE}}$), so the curvature excess and the deficit in dimension may well measure one property twice.

\subsection{Limitations and extensions}\label{sec:limitations}
    
	The absolute scale of several estimators depends on $n$, not least among them the sampled maximum $\delta$, the graph average $\bar\kappa$, and the dimension estimators at fixed $k$, scaling of which both terms of (\ref{eq:excess}) inherit. The sampling distribution of $\sigma_T$ is uncalibrated, however: it rests on $K = 10$ draws, whose distribution can depart from a Gaussian. A value of $\sigma$ near the threshold of $3$ is, therefore, weak evidence.

	The decision rule omits the spread of $T(X)$ across independent subsets of the embedding. One replicate draw puts two subsets of one condition three null standard deviations apart in $\bar\kappa$ and in $\delta/\mathrm{diameter}$, which bounds the subset spread loosely, between $0.95$ and $68$ null standard deviations at $95\%$. At the point estimate of $2.1$, a value needs about $7\sigma$ to pass the threshold; the GloVe Forman result (of $+6.2\sigma$) and both Gini results fall inside that margin and stay provisional. The dimension and curvature excesses, at $23\sigma$ to $146\sigma$, pass even there.

	The sphere null is the angular Gaussian; second moments differ from those of the cloud to which it is fit. The participation-ratio diagnostic checks one summary of the mismatch: $+1.8\sigma$ for GloVe and $+13.0\sigma$ for Word2Vec at the point level, the null more anisotropic than the data; that direction lowers the curvature excess and raises the dimension excess, so it is conservative for both findings. At the centroid level and within clusters the decreased nulls miss the diagnostic in the other direction (see Appendix~\ref{app:calibration}). The centroid $\bar\kappa$ result, therefore, rests on the rows at $k \ge 800$, and the within-cluster comparison rests on the random subsets.

	Our validation runs the dimension estimators on manifolds of intrinsic dimension at most $9$ and the participation ratio on Gaussian clouds in $\mathbb{R}^{20}$. The application regime is $\hat d \approx 26$ to $32$ on the sphere. \citet{jones2024manifold} argues that estimators built on hard neighborhoods suffer from the curse of dimensionality; both signals come from such estimators. The matched-null comparison absorbs estimator bias where the bias is common to $X$ and its null.

	The frequency-only null ran on one corpus, with one original model per configuration and three shuffles, and its attributions reach the released models only through the correspondence of Section~\ref{sec:phase3}. GloVe's shuffled-corpus models differ so far from the original's geometry that their shares are bounds. The configurations also differ in window size, subsampling, and epochs, so their contrast compares configurations as well as objectives.

	The HDBSCAN clusters as format lists and frequency bands holds for settings on each released embedding under the tied \texttt{min\_samples} design; it remains untested for $k$-means partitions, for clusterings derived from external semantic resources, and for the $205$ small Word2Vec clusters at \texttt{min\_cluster\_size} $= 5$.

	Ten or more stratified subsets per condition would measure the subset spread of every statistic, fix the effective threshold, and allow a pooled test of the sign of $\delta/\mathrm{diameter}$. On the sphere, the Fisher--Bingham distribution $p(x) \propto \exp(b^\top x + x^\top A x)$ \citep{mardiajupp2000} is the maximum-entropy law that matches the embeddings' common mean direction \citep{mu2018allbutthetopsimpleeffectivepostprocessing} and their second moment, and the sampler of \citet{kentganeibermardia2018} would draw from it. Recoloring each Gaussian draw to the sample covariance, the constrained realizations of \citet{theiler1996}, would remove the dispersion of the draws' covariances. The frequency-only null should be extended to the released models' own corpora, to several training seeds, and to configurations with matched windows and subsampling.

\section{Conclusion}\label{sec:conclusion}

	We adapted surrogate-data testing to the geometry of word embeddings as matched-null testing. On the unit sphere, GloVe and Word2Vec have lower local dimension and greater neighborhood overlap than Gaussian samples with their covariance, and $\delta/\mathrm{diameter}$ and the maximum $k$NN degree track that null. On a Wikipedia slice, shuffling the training corpus removes most of the local dimension deficit of skip-gram and GloVe, so the deficit depends on co-occurrence. Shuffled skip-gram reproduces $91\%$ of its curvature excess through hubs, with twelve times the maximum degree of the intact-corpus model; the source of that excess, therefore, is undetermined.

\bibliography{matched_null}

\appendix

\section{Every statistic against every null}\label{app:full}

	Table~\ref{tab:phase1_gap} gives observed value, mean of the covariance-matched null, raw excess, and $\sigma$ for the ten summary statistics on both embeddings on the sphere.

	\begin{table*}
		\centering
		\small
		\setlength{\tabcolsep}{3pt}
		\begin{tabular}{lrrrrrrrr}
        \toprule
         & \multicolumn{4}{c}{\textbf{GloVe, sphere}} & \multicolumn{4}{c}{\textbf{Word2Vec, sphere}} \\
        \cmidrule(lr){2-5} \cmidrule(lr){6-9}
        \textbf{Statistic} & \textbf{Observed} & \textbf{Null mean} & \textbf{Excess} & $\boldsymbol{\sigma}$ & \textbf{Observed} & \textbf{Null mean} & \textbf{Excess} & $\boldsymbol{\sigma}$ \\
        \midrule
        \tabhead{$\hat d_{\mathrm{MLE}}$} & $25.9$ & $60.6$ & $-34.7$ & $-118$ & $27.2$ & $71.3$ & $-44.1$ & $-125$ \\
        \tabhead{$\hat d_{\mathrm{corr}}$} & $17.1$ & $26.3$ & $-9.18$ & $-29$ & $31.2$ & $37.7$ & $-6.51$ & $-48$ \\
        \tabhead{$\hat d_{\mathrm{graph}}$} & $28.2$ & $72.6$ & $-44.5$ & $-85$ & $29.1$ & $84.5$ & $-55.4$ & $-57$ \\
        \tabhead{Volume growth exponent} & $15.7$ & $29.9$ & $-14.3$ & $-23$ & $29.5$ & $44.0$ & $-14.5$ & $-24$ \\
        \tabhead{$\bar\kappa$ (Ollivier)} & $0.167$ & $0.0750$ & $0.0917$ & $+45$ & $0.134$ & $0.0521$ & $0.0816$ & $+146$ \\
        \tabhead{Fraction $\kappa > 0.1$} & $0.733$ & $0.240$ & $0.494$ & $+29$ & $0.548$ & $0.0779$ & $0.470$ & $+91$ \\
        \tabhead{$\bar F$ (Forman)} & $-51.7$ & $-58.8$ & $7.11$ & $+6.2$ & $-25.9$ & $-38.6$ & $12.7$ & $+28$ \\
        \tabhead{$\delta/\mathrm{diameter}$} & $0.0954$ & $0.107$ & $-0.0112$ & $-1.3$ & $0.0859$ & $0.100$ & $-0.0142$ & $-2.0$ \\
        \tabhead{$k$NN maximum degree} & $257$ & $195$ & $62.2$ & $+1.6$ & $113$ & $82.7$ & $30.3$ & $+2.6$ \\
        \tabhead{$k$NN degree Gini} & $0.268$ & $0.255$ & $0.0124$ & $+3.9$ & $0.139$ & $0.157$ & $-0.0181$ & $-6.2$ \\
        \bottomrule
        \end{tabular}
		\caption{Point-level analysis at $n = 2000$ on the sphere. Observed value, mean of the covariance-matched null over $K$ draws, raw excess (observed minus null mean), and standardized excess $\sigma$. At large $|\sigma|$, the figure is limited by the estimate of the null standard deviation.}
		\label{tab:phase1_gap}
	\end{table*}

	Table~\ref{tab:full} lists the suite statistics and the auxiliary Forman terms at the point level with their observed values and standardized excess against the isotropic, covariance-matched, and marginal-permuted nulls, for the sphere analyses of both embeddings and the raw-row analysis of GloVe. The participation-ratio row is the second-moment diagnostic of Section~\ref{sec:nulls}; rows marked undefined have zero null variance. In the sphere analyses, $Z_{\mathrm{iso}}$ and $Z_{\mathrm{perm}}$ are re-projected to the sphere; in the raw analysis $Z_{\mathrm{iso}}$ is $\mathcal{N}(0, I_d)$ at the ambient scale, so its scale-bearing rows (diameter) are incomparable there.

	\begin{table*}
		\centering
		\scriptsize
		\setlength{\tabcolsep}{2.6pt}
		\begin{tabular}{>{\raggedright\arraybackslash}p{2.4cm}rrrrrrrrrrrr}
        \toprule
         & \multicolumn{4}{c}{\textbf{GloVe, sphere}} & \multicolumn{4}{c}{\textbf{Word2Vec, sphere}} & \multicolumn{4}{c}{\textbf{GloVe, raw}} \\
        \cmidrule(lr){2-5} \cmidrule(lr){6-9} \cmidrule(lr){10-13}
        \textbf{Statistic} & \textbf{Observed} & $\boldsymbol{\sigma_{\mathrm{iso}}}$ & $\boldsymbol{\sigma_{\mathrm{cov}}}$ & $\boldsymbol{\sigma_{\mathrm{perm}}}$ & \textbf{Observed} & $\boldsymbol{\sigma_{\mathrm{iso}}}$ & $\boldsymbol{\sigma_{\mathrm{cov}}}$ & $\boldsymbol{\sigma_{\mathrm{perm}}}$ & \textbf{Observed} & $\boldsymbol{\sigma_{\mathrm{iso}}}$ & $\boldsymbol{\sigma_{\mathrm{cov}}}$ & $\boldsymbol{\sigma_{\mathrm{perm}}}$ \\
        \midrule
        \tabhead{$\hat d_{\mathrm{MLE}}$} & $25.9$ & $-99$ & $-118$ & $-118$ & $27.2$ & $-97$ & $-125$ & $-121$ & $25.9$ & $-52$ & $-68$ & $-77$ \\
        \tabhead{$\hat d_{\mathrm{MLE}}$ local standard deviation} & $12.8$ & $-19$ & $-11$ & $-14$ & $12.0$ & $-21$ & $-12$ & $-18$ & $11.1$ & $-17$ & $-12$ & $-16$ \\
        \tabhead{$\hat d_{\mathrm{corr}}$} & $17.1$ & $-1120$ & $-29$ & $-61$ & $31.2$ & $-655$ & $-48$ & $-166$ & $11.8$ & $-73$ & $-64$ & $-92$ \\
        \tabhead{$\hat d_{\mathrm{graph}}$} & $28.2$ & $-104$ & $-85$ & $-76$ & $29.1$ & $-102$ & $-57$ & $-70$ & $27.7$ & $-43$ & $-45$ & $-48$ \\
        \tabhead{$\mathrm{dim}_{\mathrm{PCA95}}$} & $224$ & \textit{undefined} & \textit{undefined} & \textit{undefined} & $256$ & \textit{undefined} & \textit{undefined} & \textit{undefined} & $224$ & \textit{undefined} & $+17$ & \textit{undefined} \\
        \tabhead{Participation ratio} & $124$ & $-531$ & $+1.8$ & $-573$ & $178$ & $-321$ & $+13$ & $-534$ & $134$ & $-528$ & $+6.7$ & $-939$ \\
        \tabhead{$\delta/\mathrm{diameter}$} & $0.0954$ & $+3.7$ & $-1.3$ & $+1.5$ & $0.0859$ & $+0.7$ & $-2.0$ & $-1.0$ & $0.0465$ & $-10$ & $-5.3$ & $-9.8$ \\
        \tabhead{Diameter} & $1.55$ & $-11$ & $-1.8$ & $+2.4$ & $1.58$ & $-4.7$ & $-3.6$ & $+1.3$ & $23.9$ & $-27$ & $+22$ & $+43$ \\
        \tabhead{Volume growth exponent} & $15.7$ & $-140$ & $-23$ & $-18$ & $29.5$ & $-96$ & $-24$ & $-133$ & $12.5$ & $-29$ & $-27$ & $-27$ \\
        \tabhead{Volume growth $R^2$, power law} & $0.996$ & $+7.7$ & $+2.1$ & $+1.6$ & $0.997$ & $+9.8$ & $+9.3$ & $+15$ & $1.00$ & $+15$ & $+11$ & $+14$ \\
        \tabhead{Volume growth $R^2$, exponential} & $0.999$ & $+15$ & $+7.7$ & $+5.6$ & $0.998$ & $+15$ & $+16$ & $+22$ & $0.997$ & $+13$ & $+11$ & $+12$ \\
        \tabhead{$\bar\kappa$ (Ollivier)} & $0.167$ & $+235$ & $+45$ & $+32$ & $0.134$ & $+169$ & $+146$ & $+221$ & $0.219$ & $+42$ & $+40$ & $+47$ \\
        \tabhead{$\kappa$ standard deviation} & $0.0931$ & $+166$ & $+37$ & $+47$ & $0.101$ & $+184$ & $+101$ & $+205$ & $0.0795$ & $+33$ & $+50$ & $+48$ \\
        \tabhead{Fraction $\kappa > 0.1$} & $0.733$ & $+238$ & $+29$ & $+16$ & $0.548$ & $+176$ & $+91$ & $+178$ & $0.939$ & $+22$ & $+22$ & $+23$ \\
        \tabhead{$\bar F$ (Forman)} & $-51.7$ & $-256$ & $+6.2$ & $+11$ & $-25.9$ & $+47$ & $+28$ & $+29$ & $-250$ & $-11$ & $-21$ & $-13$ \\
        \tabhead{$F$ standard deviation} & $46.8$ & $+698$ & $+3.5$ & $-3.0$ & $13.0$ & $+151$ & $+0.9$ & $+1.8$ & $398$ & $+12$ & $+21$ & $+10$ \\
        \tabhead{Fraction $F < -10$} & $0.933$ & \textit{undefined} & $-5343$ & \textit{undefined} & $0.915$ & \textit{undefined} & \textit{undefined} & \textit{undefined} & $0.941$ & $-3688$ & $-1335$ & $-4953$ \\
        \tabhead{$k$NN maximum degree} & $257$ & $+184$ & $+1.6$ & $-2.2$ & $113$ & $+67$ & $+2.6$ & $+8.5$ & $1553$ & $+8.9$ & $+17$ & $+5.6$ \\
        \tabhead{$k$NN degree Gini} & $0.268$ & $+179$ & $+3.9$ & $-19$ & $0.139$ & $+63$ & $-6.2$ & $-5.5$ & $0.378$ & $+4.5$ & $+11$ & $+11$ \\
        \midrule
        \tabhead{Mean degree (auxiliary)} & $24.9$ & $+240$ & $-0.9$ & $-20$ & $19.9$ & $+87$ & $-4.2$ & $-2.5$ & $27.4$ & $+6.4$ & $+11$ & $+13$ \\
        \tabhead{Forman degree term (auxiliary)} & $77.7$ & $+542$ & $+5.2$ & $-6.4$ & $44.0$ & $+115$ & $-3.1$ & $-1.9$ & $283$ & $+13$ & $+22$ & $+14$ \\
        \tabhead{Forman triangle term (auxiliary)} & $7.34$ & $+947$ & $+45$ & $+25$ & $4.72$ & $+593$ & $+105$ & $+298$ & $9.78$ & $+31$ & $+34$ & $+31$ \\
        \bottomrule
        \end{tabular}
		\caption{Observed value and$\sigma$ against each null; $K = 10$ draws per null.}
		\label{tab:full}
	\end{table*}

\section{Null calibration}\label{app:calibration}

	The Ledoit--Wolf null of Section~\ref{sec:nulls} is more isotropic than an anisotropic generator. On structureless anisotropic Gaussian unit vectors at $n \in \{50, 100, 200\}$ and $d = 300$, the lessening by itself produces $\sigma_{\bar\kappa}$ between $+2.4$ and $+5.1$ and $\sigma_{\hat d_{\mathrm{MLE}}}$ between $-2.9$ and $-4.6$; on isotropic unit vectors at the same $(n, d)$, $|\sigma| \le 2.6$. The per-cluster nulls of Section~\ref{sec:within} miss the participation-ratio diagnostic by $5$ to $47\sigma$.

	Table~\ref{tab:centroid_full} gives every $k$-means clustering of the centroid test. For $n_{\mathrm{cent}}$ up to $400$, the participation ratio of the centroid cloud sits $6$ to $10\sigma$ below its shrunk null: the null is more isotropic than the centroids. The standardized $\bar\kappa$ excesses of those rows, $+3.8$ to $+41$, have the sign of the shrinkage artifact and stay confounded with it. For $\delta$ the artifact runs toward positive values ($+0.6$ to $+3.9$ on the anisotropic calibration), so it can mask a deficit. Across all twelve clusterings $\sigma_\delta$ lies in $[-1.8, +2.4]$, inside $\pm 2.5$, and the six values at $k \ge 400$ take both signs.

	\begin{table*}
		\centering
		\small
		\setlength{\tabcolsep}{4pt}
		\begin{tabular}{lrlrrrrrrr}
        \toprule
        \textbf{Clustering} & $\boldsymbol{n_{\mathrm{cent}}}$ & \textbf{Null} & $\boldsymbol{\delta_{\mathrm{rel}}}$ & $\boldsymbol{\sigma_{\delta}}$ & $\boldsymbol{\bar\kappa}$ & $\boldsymbol{\bar\kappa}$ \textbf{null} & $\boldsymbol{\sigma_{\bar\kappa}}$ & $\boldsymbol{\sigma_{\hat d}}$ & $\boldsymbol{\sigma_{P}}$ \\
        \midrule
        \tabhead{GloVe $k$-means $k=50$} & 50 & shrunk & $0.129$ & $+0.1$ & $0.300$ & $0.270$ & $+3.8$ & $-12$ & $-9.0$ \\
        \tabhead{GloVe $k$-means $k=100$} & 100 & shrunk & $0.150$ & $+2.4$ & $0.241$ & $0.200$ & $+9.0$ & $-21$ & $-6.9$ \\
        \tabhead{GloVe $k$-means $k=200$} & 186 & shrunk & $0.129$ & $-0.1$ & $0.215$ & $0.160$ & $+21$ & $-32$ & $-7.8$ \\
        \tabhead{GloVe $k$-means $k=400$} & 397 & shrunk & $0.127$ & $+0.7$ & $0.190$ & $0.126$ & $+41$ & $-58$ & $-6.9$ \\
        \tabhead{GloVe $k$-means $k=800$} & 737 & empirical & $0.127$ & $-0.8$ & $0.178$ & $0.102$ & $+69$ & $-42$ & $+2.2$ \\
        \tabhead{GloVe $k$-means $k=1000$} & 851 & empirical & $0.108$ & $-1.8$ & $0.175$ & $0.101$ & $+40$ & $-46$ & $+1.9$ \\
        \tabhead{Word2Vec $k$-means $k=50$} & 50 & shrunk & $0.137$ & $+2.1$ & $0.288$ & $0.256$ & $+6.3$ & $-13$ & $-6.8$ \\
        \tabhead{Word2Vec $k$-means $k=100$} & 100 & shrunk & $0.135$ & $+1.4$ & $0.241$ & $0.185$ & $+13$ & $-17$ & $-6.4$ \\
        \tabhead{Word2Vec $k$-means $k=200$} & 194 & shrunk & $0.132$ & $+1.9$ & $0.210$ & $0.140$ & $+23$ & $-29$ & $-9.7$ \\
        \tabhead{Word2Vec $k$-means $k=400$} & 400 & shrunk & $0.119$ & $-0.2$ & $0.185$ & $0.107$ & $+41$ & $-46$ & $-9.6$ \\
        \tabhead{Word2Vec $k$-means $k=800$} & 771 & empirical & $0.133$ & $+1.4$ & $0.164$ & $0.0870$ & $+48$ & $-72$ & $+5.1$ \\
        \tabhead{Word2Vec $k$-means $k=1000$} & 909 & empirical & $0.119$ & $-0.4$ & $0.161$ & $0.0826$ & $+53$ & $-60$ & $+5.1$ \\
        \bottomrule
        \end{tabular}
        
		\caption{$n_{\mathrm{cent}}$ counts centroids of clusters with at least $50$ members at $k \le 200$ and $10$ at $k \ge 400$; the covariance-matched null is Ledoit--Wolf shrunk when $n_{\mathrm{cent}} < 2d$ and empirical otherwise. Observed $\delta/\mathrm{diameter}$ and $\bar\kappa$ with the null mean of $\bar\kappa$; $\sigma_{\delta}$, $\sigma_{\bar\kappa}$, and $\sigma_{\hat d}$ are the standardized excesses of $\delta/\mathrm{diameter}$, $\bar\kappa$, and $\hat d_{\mathrm{MLE}}$, and $\sigma_{P}$ that of the participation ratio $P$, the second-moment diagnostic of Section~\ref{sec:nulls}.}
		\label{tab:centroid_full}
	\end{table*}

\section{Clustering details}\label{app:clusters}

	On a structureless anisotropic Gaussian cloud projected to the sphere ($d = 300$, spectrum $\lambda_i \propto i^{-0.6}$, participation ratio $125$ after projection, close to GloVe's $124$), the estimators give $\bar\kappa = 0.067$, $\hat d_{\mathrm{MLE}} = 63.8$, and $\mathrm{dim}_{\mathrm{PCA95}} = 251$ at $n = 2000$; $0.099$, $54.8$, and $205$ at $n = 500$; $0.121$, $49.1$, and $154$ at $n = 250$; and $0.231$, $31.7$, and $43$ at $n = 50$. Table~\ref{tab:within} shows medians across clusters of the within-cluster statistics and of their random-subset controls.

	\begin{table}
		\centering
		\footnotesize
		\setlength{\tabcolsep}{3.5pt}
		\begin{tabular}{lrrrr}
        \toprule
         & \multicolumn{2}{c}{\tabhead{$k$-means $k=100$}} & \multicolumn{2}{c}{\textbf{HDBSCAN clusters}} \\
        \cmidrule(lr){2-3} \cmidrule(lr){4-5}
        \textbf{Statistic} & \textbf{Cluster} & \textbf{Control} & \textbf{Cluster} & \textbf{Control} \\
        \midrule
        \multicolumn{5}{l}{\textbf{GloVe}} \\
        \tabhead{$\bar\kappa$} & $0.205$ & $0.199$ & $0.336$ & $0.274$ \\
        \tabhead{$\hat d_{\mathrm{MLE}}$} & $24.7$ & $26.9$ & $9.71$ & $25.1$ \\
        \tabhead{$\delta/\mathrm{diameter}$} & $0.111$ & $0.103$ & $0.129$ & $0.101$ \\
        \tabhead{$\mathrm{dim}_{\mathrm{PCA95}}$} & $146$ & $154$ & $40$ & $44$ \\
        \multicolumn{5}{l}{\textbf{Word2Vec}} \\
        \tabhead{$\bar\kappa$} & $0.164$ & $0.146$ & $0.298$ & $0.151$ \\
        \tabhead{$\hat d_{\mathrm{MLE}}$} & $28.1$ & $31.7$ & $17.0$ & $32.2$ \\
        \tabhead{$\delta/\mathrm{diameter}$} & $0.106$ & $0.0919$ & $0.115$ & $0.0929$ \\
        \tabhead{$\mathrm{dim}_{\mathrm{PCA95}}$} & $158$ & $175$ & $103$ & $138$ \\
        \bottomrule
        \end{tabular}
		\caption{Within-cluster geometry at $k$-means $k = 100$ and for the HDBSCAN clusters: medians across the analyzed clusters of the observed statistic and of the same statistic on random subsets of the cloud at the cluster sizes (control). HDBSCAN uses \texttt{min\_cluster\_size} $= 50$ for GloVe (three clusters of $863$, $54$, and $51$ points) and \texttt{min\_cluster\_size} $= 100$ for Word2Vec (three clusters of $502$, $204$, and $186$ points); the two smallest GloVe clusters are below the regime of the $k = 20$ dimension estimator. Per-cluster covariance-matched nulls appear in Figure~\ref{fig:stat_vs_n}; they are shrunk below $2d$ points and miss the participation ratio by $5$ to $47\sigma$, so their $\sigma$ values are biased toward positive curvature excess; the figure shows them, and the table omits them.}
		\label{tab:within}
	\end{table}

\begin{figure}
		\centering
		\includegraphics[width=0.6\textwidth]{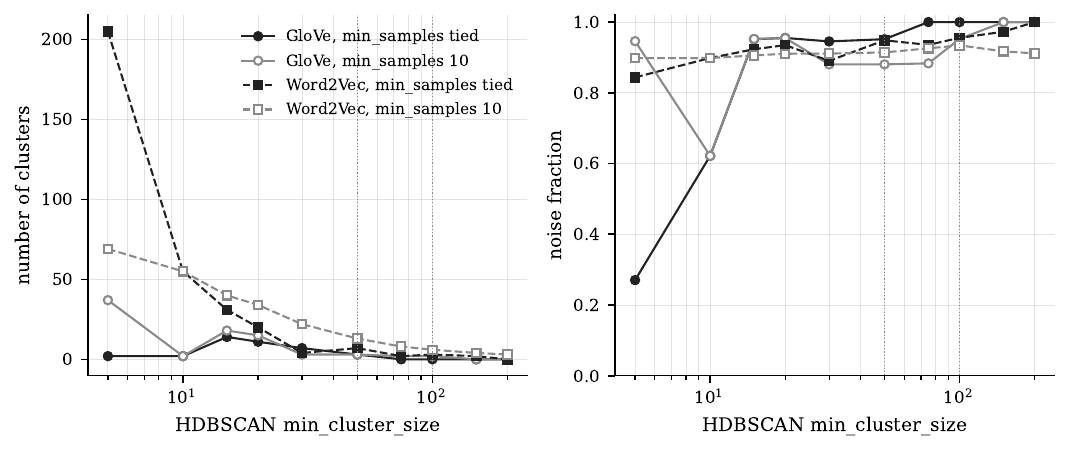}
		\caption{HDBSCAN cluster count (left) and noise fraction (right) against \texttt{min\_cluster\_size} on a log axis, for GloVe (solid, circles) and Word2Vec (dashed, squares), with \texttt{min\_samples} tied (filled markers) and fixed at $10$ (open markers). Dotted lines mark \texttt{min\_cluster\_size} values of $50$ and $100$.}
		\label{fig:mcs_sweep}
	\end{figure}

	If the clusters are a fixed feature of the data, then the cluster count should be approximately step-shaped in \texttt{min\_cluster\_size}. Table~\ref{tab:mcs_sweep} and Figure~\ref{fig:mcs_sweep} show: with \texttt{min\_samples} tied to \texttt{min\_cluster\_size}, the library default and the setting of Section~\ref{sec:phase2a}, each step changes the density estimate as well as the size cutoff; the non-monotone Word2Vec counts ($4 \to 7$ between \texttt{min\_cluster\_size} $= 30$ and $50$, $2 \to 3$ between $75$ and $100$) and a $1{,}976$-point cluster at \texttt{min\_cluster\_size} $= 30$ come from this. With \texttt{min\_samples} fixed at $10$ the core distances and the single-linkage hierarchy are fixed, and \texttt{min\_cluster\_size} acts only on condensation and excess-of-mass extraction; the count can still change in either direction, because extraction may select children in place of a parent, but a change is then a property of the extraction stage on one hierarchy, separate from any change in the density estimate.

	\begin{table*}
		\centering
		\small
		\begin{tabular}{lrrrrrrrr}
        \toprule
         & \multicolumn{4}{c}{\textbf{GloVe}} & \multicolumn{4}{c}{\textbf{Word2Vec}} \\
        \cmidrule(lr){2-5} \cmidrule(lr){6-9}
         & \multicolumn{2}{c}{\textbf{Tied design}} & \multicolumn{2}{c}{\textbf{Fixed design}} & \multicolumn{2}{c}{\textbf{Tied design}} & \multicolumn{2}{c}{\textbf{Fixed design}} \\
        \cmidrule(lr){2-3} \cmidrule(lr){4-5} \cmidrule(lr){6-7} \cmidrule(lr){8-9}
        \textbf{Minimum cluster size} & \textbf{Clusters} & \textbf{Noise} & \textbf{Clusters} & \textbf{Noise} & \textbf{Clusters} & \textbf{Noise} & \textbf{Clusters} & \textbf{Noise} \\
        \midrule
        \tabhead{$5$} & $2$ & $0.27$ & $37$ & $0.95$ & $205$ & $0.84$ & $69$ & $0.90$ \\
        \tabhead{$10$} & $2$ & $0.62$ & $2$ & $0.62$ & $55$ & $0.90$ & $55$ & $0.90$ \\
        \tabhead{$15$} & $14$ & $0.95$ & $18$ & $0.95$ & $31$ & $0.92$ & $40$ & $0.91$ \\
        \tabhead{$20$} & $11$ & $0.96$ & $15$ & $0.95$ & $20$ & $0.93$ & $34$ & $0.91$ \\
        \tabhead{$30$} & $7$ & $0.95$ & $3$ & $0.88$ & $4$ & $0.89$ & $22$ & $0.91$ \\
        \tabhead{$50$} & $3$ & $0.95$ & $3$ & $0.88$ & $7$ & $0.95$ & $13$ & $0.91$ \\
        \tabhead{$75$} & $0$ & $1.00$ & $2$ & $0.88$ & $2$ & $0.94$ & $8$ & $0.93$ \\
        \tabhead{$100$} & $0$ & $1.00$ & $2$ & $0.95$ & $3$ & $0.96$ & $6$ & $0.93$ \\
        \tabhead{$150$} & $0$ & $1.00$ & $0$ & $1.00$ & $2$ & $0.97$ & $4$ & $0.92$ \\
        \tabhead{$200$} & $0$ & $1.00$ & $0$ & $1.00$ & $0$ & $1.00$ & $3$ & $0.91$ \\
        \bottomrule
        \end{tabular}
		\caption{HDBSCAN cluster count and noise fraction against \texttt{min\_cluster\_size} on the $n = 20{,}000$ subset, with \texttt{min\_samples} tied to \texttt{min\_cluster\_size} (tied design) and fixed at $10$ (fixed design).}
		\label{tab:mcs_sweep}
	\end{table*}

	Under the tied design, both cluster counts move with the threshold: at \texttt{min\_cluster\_size} $\le 10$ GloVe has one macro-cluster holding $73\%$ of the points at \texttt{min\_cluster\_size} $= 5$ ($38\%$ at $10$) beside a second cluster of $7$ and $10$ points; the count peaks at $14$ around $15$, falls to three clusters at $50$, and is zero from $75$. Word2Vec falls from $205$ clusters at \texttt{min\_cluster\_size} $= 5$ (uninspected) to a few clusters of several hundred points from $75$ to $150$, and to zero at $200$. Under the fixed design, Word2Vec declines through $69$, $55$, $40$, $34$, $22$, $13$, $8$, $6$, $4$, and $3$ clusters with the noise fraction between $0.90$ and $0.93$ throughout, a smooth decline from $69$ clusters to $3$. GloVe under the fixed design is non-monotone ($37$, $2$, $18$, $15$, $3$, $3$, $2$, $2$, $0$, $0$): at \texttt{min\_cluster\_size} $= 10$ extraction selects the $7{,}555$-point macro-cluster, at $15$ and $20$ it selects its children, and from $30$ to $75$ a single $2{,}249$-point cluster, $11\%$ of the points, holds the high-frequency region at this density setting. 
    
    Word2Vec, trained on Google News, contains many small clusters detectable at low density thresholds, and GloVe, trained on Dolma, has one broad low-density mass that resolves into a handful of clusters over a narrow band of thresholds. Corpus, training configuration, casing, and phrase tokens all differ between them. Under both designs, the cluster count depends on the threshold, and under the fixed design the extraction stage changes it by itself.

	Table~\ref{tab:diffuse} gives the values behind Figure~\ref{fig:diffuse_preservation}.

\begin{table*}
	\centering
	\small
	\setlength{\tabcolsep}{4pt}
	\begin{tabular}{lrrrrrr}
		\toprule
		\textbf{Condition} & \textbf{Minimum cluster size} & $\boldsymbol{\bar\kappa}$ & \textbf{Null mean} & \textbf{Excess} & \textbf{Ratio} & $\boldsymbol{\sigma}$ \\
		\midrule
		\tabhead{GloVe, full subset} & & $0.167$ & $0.0750$ & $0.0917$ & $2.22$ & $+45$ \\
		\tabhead{GloVe, 3 clusters removed (968 points)} & $50$ & $0.137$ & $0.0698$ & $0.0667$ & $1.95$ & $+54$ \\
		\tabhead{GloVe, numeric clusters removed (105 points)} & $50$ & $0.164$ & $0.0755$ & $0.0881$ & $2.17$ & $+77$ \\
		\tabhead{GloVe, replicate draw} & $100$ & $0.161$ & $0.0754$ & $0.0851$ & $2.13$ & $+71$ \\
		\midrule
		\tabhead{Word2Vec, full subset} & & $0.134$ & $0.0521$ & $0.0816$ & $2.57$ & $+146$ \\
		\tabhead{Word2Vec, 7 clusters removed (1{,}032 points)} & $50$ & $0.126$ & $0.0520$ & $0.0743$ & $2.43$ & $+122$ \\
		\tabhead{Word2Vec, 3 clusters removed (892 points)} & $100$ & $0.124$ & $0.0516$ & $0.0723$ & $2.40$ & $+124$ \\
		\bottomrule
	\end{tabular}
	\caption{Ollivier $\bar\kappa$ on the full subset and after cluster removal: observed value, mean of the covariance-matched null, raw excess, ratio to the null, and $\sigma$. The minimum cluster size is the HDBSCAN setting whose clusters were removed; at $100$, HDBSCAN every GloVe as noise, row is a replicate draw of the full condition. Cluster-removed draws are frequency stratified over the remainder; $K = 10$.}
	\label{tab:diffuse}
\end{table*}

\section{Retraining details}\label{app:retraining}

	Tables~\ref{tab:shuffled} and~\ref{tab:shuffled_glove} show the full contrast of Section~\ref{sec:phase3} on the sphere, and Tables~\ref{tab:shuffled_raw} and~\ref{tab:shuffled_glove_raw} repeat it on the raw rows. On the raw rows of its $n = 2000$ subset, the Wikipedia skip-gram model has norm coefficient of variation $0.13$ and Euclidean--cosine rank correlation $0.43$, and the Wikipedia GloVe model $0.35$ and $0.22$; the shuffled skip-gram models have $0.15$ and $0.92$ with a Ledoit--Wolf weight of $0.96$ against the original's $0.12$, and the shuffled GloVe models $0.70$ and $0.48$ with a weight of $0.009$.

	\begin{table*}
		\centering
		\small
		\setlength{\tabcolsep}{4pt}
		\begin{tabular}{lrrrrrrr}
        \toprule
         & \multicolumn{2}{c}{\textbf{Original model}} & \multicolumn{3}{c}{\textbf{Shuffled-corpus models}} & & \\
        \cmidrule(lr){2-3} \cmidrule(lr){4-6}
        \textbf{Statistic} & \textbf{Value} & \textbf{Excess} & \textbf{Mean} & \textbf{Standard deviation} & \textbf{Excess} & \textbf{Share} & $\boldsymbol{\sigma_{\mathrm{shuffle}}}$ \\
        \midrule
        \tabhead{$\hat d_{\mathrm{MLE}}$} & $27.2$ & $-33.8$ & $71.1$ & $0.948$ & $-4.56$ & $0.13$ & $-46$ \\
        \tabhead{$\hat d_{\mathrm{corr}}$} & $24.6$ & $-7.26$ & $22.7$ & $0.0490$ & $-9.35$ & $1.29$ & $+39$ \\
        \tabhead{$\hat d_{\mathrm{graph}}$} & $31.2$ & $-41.6$ & $86.4$ & $0.962$ & $-3.85$ & $0.09$ & $-57$ \\
        \tabhead{Volume growth exponent} & $23.1$ & $-14.2$ & $25.1$ & $0.370$ & $-12.5$ & $0.88$ & $-5.5$ \\
        \tabhead{$\bar\kappa$ (Ollivier)} & $0.158$ & $0.101$ & $0.175$ & $0.00730$ & $0.0924$ & $0.91$ & $-2.2$ \\
        \tabhead{Fraction $\kappa > 0.1$} & $0.647$ & $0.519$ & $0.958$ & $0.0145$ & $0.677$ & $1.30$ & $-21$ \\
        \tabhead{$\bar F$ (Forman)} & $-27.5$ & $10.3$ & $-269$ & $18.3$ & $-165$ & $-15.92$ & $+13$ \\
        \tabhead{Forman triangle term (auxiliary)} & $6.12$ & $4.73$ & $7.50$ & $0.518$ & $4.93$ & $1.04$ & $-2.7$ \\
        \tabhead{Forman degree term (auxiliary)} & $49.9$ & $3.84$ & $296$ & $19.9$ & $180$ & $46.74$ & $-12$ \\
        \tabhead{$\delta/\mathrm{diameter}$} & $0.102$ & $-0.00562$ & $0.0820$ & $0.00576$ & $-0.000294$ & $0.05$ & $+3.5$ \\
        \tabhead{$k$NN maximum degree} & $74$ & $0.200$ & $900$ & $106$ & $456$ & $2282.50$ & $-7.8$ \\
        \tabhead{$k$NN degree Gini} & $0.184$ & $0.0209$ & $0.425$ & $0.00384$ & $0.0803$ & $3.83$ & $-63$ \\
        \tabhead{Participation ratio} & $127$ & $4.83$ & $256$ & $0.223$ & $26.7$ & $5.52$ & $-577$ \\
        \bottomrule
        \end{tabular}
		\caption{Skip-gram on the original and on the shuffled Wikipedia slice, sphere geometry.}
		\label{tab:shuffled}
	\end{table*}

	\begin{table*}
		\centering
		\small
		\setlength{\tabcolsep}{4pt}
		\begin{tabular}{lrrrrrrr}
        \toprule
         & \multicolumn{2}{c}{\textbf{Original model}} & \multicolumn{3}{c}{\textbf{Shuffled-corpus models}} & & \\
        \cmidrule(lr){2-3} \cmidrule(lr){4-6}
        \textbf{Statistic} & \textbf{Value} & \textbf{Excess} & \textbf{Mean} & \textbf{Standard deviation} & \textbf{Excess} & \textbf{Share} & $\boldsymbol{\sigma_{\mathrm{shuffle}}}$ \\
        \midrule
        \tabhead{$\hat d_{\mathrm{MLE}}$} & $32.1$ & $-21.6$ & $42.2$ & $0.248$ & $4.03$ & $-0.19$ & $-41$ \\
        \tabhead{$\hat d_{\mathrm{corr}}$} & $24.4$ & $-3.73$ & $6.04$ & $0.0444$ & $-4.42$ & $1.19$ & $+414$ \\
        \tabhead{$\hat d_{\mathrm{graph}}$} & $35.2$ & $-29.0$ & $51.7$ & $0.366$ & $4.40$ & $-0.15$ & $-45$ \\
        \tabhead{Volume growth exponent} & $24.3$ & $-7.30$ & $5.94$ & $0.0636$ & $-6.31$ & $0.87$ & $+289$ \\
        \tabhead{$\bar\kappa$ (Ollivier)} & $0.105$ & $0.0427$ & $0.118$ & $0.00196$ & $-0.0373$ & $-0.87$ & $-6.5$ \\
        \tabhead{Fraction $\kappa > 0.1$} & $0.449$ & $0.271$ & $0.547$ & $0.00755$ & $-0.248$ & $-0.92$ & $-13$ \\
        \tabhead{$\bar F$ (Forman)} & $-33.0$ & $6.84$ & $-53.8$ & $0.887$ & $15.4$ & $2.25$ & $+24$ \\
        \tabhead{Forman triangle term (auxiliary)} & $4.07$ & $2.21$ & $4.75$ & $0.107$ & $-1.44$ & $-0.65$ & $-6.4$ \\
        \tabhead{Forman degree term (auxiliary)} & $49.2$ & $-0.199$ & $72.1$ & $1.20$ & $-19.7$ & $98.85$ & $-19$ \\
        \tabhead{$\delta/\mathrm{diameter}$} & $0.111$ & $-0.0102$ & $0.135$ & $0.00911$ & $-0.00896$ & $0.88$ & $-2.7$ \\
        \tabhead{$k$NN maximum degree} & $78$ & $-20.1$ & $186$ & $11.4$ & $-58.5$ & $2.91$ & $-9.5$ \\
        \tabhead{$k$NN degree Gini} & $0.177$ & $-0.00203$ & $0.268$ & $0.00223$ & $-0.0515$ & $25.45$ & $-40$ \\
        \tabhead{Participation ratio} & $95.8$ & $-0.181$ & $11.1$ & $0.0592$ & $-4.41$ & $24.32$ & $+1430$ \\
        \bottomrule
        \end{tabular}
		\caption{GloVe on the original and on the shuffled Wikipedia slice, sphere geometry, laid out as Table~\ref{tab:shuffled}. The shuffled models collapse to a participation ratio of $11$.}
		\label{tab:shuffled_glove}
	\end{table*}

	\begin{table*}
		\centering
		\small
		\setlength{\tabcolsep}{4pt}
		\begin{tabular}{lrrrrrrr}
        \toprule
         & \multicolumn{2}{c}{\textbf{Original model}} & \multicolumn{3}{c}{\textbf{Shuffled-corpus models}} & & \\
        \cmidrule(lr){2-3} \cmidrule(lr){4-6}
        \textbf{Statistic} & \textbf{Value} & \textbf{Excess} & \textbf{Mean} & \textbf{Standard deviation} & \textbf{Excess} & \textbf{Share} & $\boldsymbol{\sigma_{\mathrm{shuffle}}}$ \\
        \midrule
        \tabhead{$\hat d_{\mathrm{MLE}}$} & $28.3$ & $-30.5$ & $52.9$ & $0.184$ & $-22.1$ & $0.73$ & $-134$ \\
        \tabhead{$\hat d_{\mathrm{corr}}$} & $13.6$ & $-12.2$ & $8.18$ & $0.0666$ & $-26.1$ & $2.14$ & $+81$ \\
        \tabhead{$\hat d_{\mathrm{graph}}$} & $32.9$ & $-37.8$ & $72.5$ & $1.41$ & $-16.9$ & $0.45$ & $-28$ \\
        \tabhead{Volume growth exponent} & $15.7$ & $-14.9$ & $10.0$ & $0.131$ & $-30.7$ & $2.05$ & $+44$ \\
        \tabhead{$\bar\kappa$ (Ollivier)} & $0.199$ & $0.119$ & $0.350$ & $0.00335$ & $0.277$ & $2.34$ & $-45$ \\
        \tabhead{Fraction $\kappa > 0.1$} & $0.876$ & $0.597$ & $1$ & $0$ & $0.801$ & $1.34$ & \textit{undefined} \\
        \tabhead{$\bar F$ (Forman)} & $-116$ & $-31.6$ & $-797$ & $6.48$ & $-693$ & $21.91$ & $+105$ \\
        \tabhead{Forman triangle term (auxiliary)} & $8.85$ & $6.08$ & $15.4$ & $0.119$ & $13.3$ & $2.18$ & $-55$ \\
        \tabhead{Forman degree term (auxiliary)} & $147$ & $49.9$ & $847$ & $6.65$ & $733$ & $14.69$ & $-105$ \\
        \tabhead{$\delta/\mathrm{diameter}$} & $0.0875$ & $-0.0104$ & $0.0703$ & $0.00527$ & $-0.0127$ & $1.22$ & $+3.3$ \\
        \tabhead{$k$NN maximum degree} & $767$ & $337$ & $1602$ & $107$ & $1151$ & $3.41$ & $-7.8$ \\
        \tabhead{$k$NN degree Gini} & $0.336$ & $0.0263$ & $0.483$ & $0.000744$ & $0.145$ & $5.53$ & $-197$ \\
        \tabhead{Participation ratio} & $127$ & $7.71$ & $253$ & $0.224$ & $28.8$ & $3.73$ & $-565$ \\
        \bottomrule
        \end{tabular}
		\caption{Skip-gram, raw rows.}
		\label{tab:shuffled_raw}
	\end{table*}

	\begin{table*}
		\centering
		\small
		\setlength{\tabcolsep}{4pt}
		\begin{tabular}{lrrrrrrr}
        \toprule
         & \multicolumn{2}{c}{\textbf{Original model}} & \multicolumn{3}{c}{\textbf{Shuffled-corpus models}} & & \\
        \cmidrule(lr){2-3} \cmidrule(lr){4-6}
        \textbf{Statistic} & \textbf{Value} & \textbf{Excess} & \textbf{Mean} & \textbf{Standard deviation} & \textbf{Excess} & \textbf{Share} & $\boldsymbol{\sigma_{\mathrm{shuffle}}}$ \\
        \midrule
        \tabhead{$\hat d_{\mathrm{MLE}}$} & $39.5$ & $-13.8$ & $39.4$ & $0.150$ & $3.57$ & $-0.26$ & $+1.0$ \\
        \tabhead{$\hat d_{\mathrm{corr}}$} & $5.38$ & $-18.1$ & $8.59$ & $0.128$ & $2.11$ & $-0.12$ & $-25$ \\
        \tabhead{$\hat d_{\mathrm{graph}}$} & $46.8$ & $-17.2$ & $48.5$ & $0.505$ & $2.84$ & $-0.16$ & $-3.3$ \\
        \tabhead{Volume growth exponent} & $6.28$ & $-22.2$ & $13.0$ & $0.294$ & $1.73$ & $-0.08$ & $-23$ \\
        \tabhead{$\bar\kappa$ (Ollivier)} & $0.278$ & $0.198$ & $0.121$ & $0.00219$ & $-0.0161$ & $-0.08$ & $+72$ \\
        \tabhead{Fraction $\kappa > 0.1$} & $0.991$ & $0.712$ & $0.569$ & $0.0333$ & $-0.120$ & $-0.17$ & $+13$ \\
        \tabhead{$\bar F$ (Forman)} & $-158$ & $-81.6$ & $-53.4$ & $2.10$ & $-16.1$ & $0.20$ & $-50$ \\
        \tabhead{Forman triangle term (auxiliary)} & $9.33$ & $6.65$ & $5.01$ & $0.119$ & $-0.530$ & $-0.08$ & $+36$ \\
        \tabhead{Forman degree term (auxiliary)} & $190$ & $102$ & $72.4$ & $2.44$ & $14.5$ & $0.14$ & $+48$ \\
        \tabhead{$\delta/\mathrm{diameter}$} & $0.0801$ & $-0.0233$ & $0.0492$ & $0.00216$ & $-0.0256$ & $1.10$ & $+14$ \\
        \tabhead{$k$NN maximum degree} & $676$ & $353$ & $193$ & $37.2$ & $46.2$ & $0.13$ & $+13$ \\
        \tabhead{$k$NN degree Gini} & $0.414$ & $0.112$ & $0.272$ & $0.00382$ & $0.0505$ & $0.45$ & $+37$ \\
        \tabhead{Participation ratio} & $96.9$ & $4.25$ & $2.34$ & $0.0132$ & $-0.00438$ & $-0.00$ & $+7161$ \\
        \bottomrule
        \end{tabular}
		\caption{GloVe, raw rows.}
		\label{tab:shuffled_glove_raw}
	\end{table*}

\end{document}